\documentclass[11pt]{article}

\usepackage[T1]{fontenc}
\usepackage[utf8]{inputenc}
\usepackage{lmodern}
\usepackage{microtype}
\usepackage[a4paper,margin=1in]{geometry}
\usepackage{amsmath,amssymb,amsthm}
\usepackage{mathtools}
\usepackage{booktabs}
\usepackage{enumitem}
\usepackage{xcolor}
\usepackage[round,authoryear]{natbib}
\usepackage{hyperref}
\usepackage[nameinlink,noabbrev]{cleveref}
\usepackage{listings}

\lstdefinelanguage{Solidity}{keywords={function, public, external, onlyOwner, require, emit,
uint, uint256, address, bool, returns, event, contract, mappings, msg, sender, value},
keywordstyle=\color{blue!70!black}\bfseries,
ndkeywords={true, false, this},
ndkeywordstyle=\color{purple}\bfseries,
identifierstyle=\color{black},
sensitive=true,
comment=[l]{//},
commentstyle=\color{gray}\ttfamily,
stringstyle=\color{teal}\ttfamily,
morestring=[b]'',
morestring=[b]""
}
\title{From Dependency to Compositionality:\\
A Neurosymbolic Lifting of LLM Outputs via Combinatory Categorial Grammar}
\author{Remo Pareschi\\
\small STAKE Lab, University of Molise, Campobasso, Italy\\
\small Corresponding author: \texttt{remo.pareschi@unimol.it}\\
\small ORCID: 0000-0002-4912-582X}
\date{}

\newcommand{\catS}{S}
\newcommand{\catNP}{NP}
\newcommand{\catN}{N}

\begin{document}
\maketitle

\begin{abstract}
Large language models (LLMs) generate fluent text by incrementally predicting the next token from a prefix. Critics in the generative tradition argue that such systems lack genuine grammar; influential replies from the dependency-grammar perspective hold that LLM behavior is well described by local head--dependent structure built word by word. We argue that a sharper observation has been overlooked: the prefix-driven, type-completing dynamics of autoregressive generation align closely with the incremental processing model that Combinatory Categorial Grammar (CCG) was originally designed to support. On this basis we propose a neurosymbolic framework in which LLM outputs are \emph{lifted} into typed compositional derivations --- not claiming that LLMs implement CCG internally, but that their outputs admit a principled, incremental, and auditable CCG reconstruction. Two consequences follow. First, through the Curry--Howard correspondence the lifting extends beyond natural language to the formal languages LLMs also produce --- programming languages such as Solidity, description-logic and query languages such as OWL and SQL, and the autoformalization of mathematics into proof assistants such as Lean --- with the type system varying and the architecture held fixed. Second, the lifting supports two layers of checking: a compositional layer that catches structural failures directly, and a content layer that checks the lifted structure against external knowledge sources, enabling the earliest possible flagging of hallucinated content. A current dispute over the Lean formalization of an announced Navier--Stokes proof is read as this distinction operating in the wild: compilation certifies composition, not correspondence. A feasibility study on 2,922 sentences generated by four LLMs in five registers shows that the lifting covers 97.7\% of them, that normal-form and incremental derivations receive identical logical forms wherever both are well-formed, and that most checkable predications are available before the sentence is complete. The account thereby requires of a producer not cognition but a prefix-driven generative profile. We close with a sketch of synchronous LLM--CCG coupling as one direction the framework opens.
\end{abstract}


\section{Introduction}
\label{sec:intro}

The success of large language models (LLMs) has reopened, in an unexpected form, a number of classical disputes of theoretical linguistics. The fluency and broad syntactic competence of these systems have led many observers to regard them as evidence that grammatical regularities can emerge from large-scale statistical learning. Critics in the generative tradition have replied that such systems remain fundamentally grammarless: that they manipulate word sequences without possessing structured principles of the kind that underwrite genuine linguistic knowledge.

A recent and influential reply to this critique has come from psycholinguistic work in the dependency-grammar tradition, most prominently associated with Edward Gibson. On this view, the surface behavior of LLMs is well described by local head--dependent relations that are formed incrementally as words are processed, and many of the empirical regularities observed both in human comprehension and in LLM behavior follow from a single principle of dependency locality \citep{gibson1998,futrell2020,gibson2025}. This shifts the debate away from the binary question of whether LLMs ``have grammar,'' toward the more precise question of whether their observable behavior admits a recoverable grammatical organization. In that respect, dependency grammar offers a compelling \emph{first} layer of interpretation.

The present paper begins with the observation that, while this dependency-based reply is well taken, it leaves an even more interesting convergence unstated. Autoregressive LLMs operate prefix by prefix: at each step they maintain a representation of what has been produced so far and predict what could come next. Combinatory Categorial Grammar (CCG), as developed by Mark Steedman and colleagues, was designed precisely to support such a left-to-right, prefix-driven, type-completing notion of processing \citep{ades-steedman-1982,steedman2000,steedman2011}. In CCG, partial expressions receive principled syntactic and semantic types --- ``John loves'' is not an ill-formed fragment but a category $\catS/\catNP$ with an associated lambda term --- and well-defined combinatory rules govern how those types extend as new material arrives. The same incremental, predictive, prefix-completing dynamic that LLMs realize statistically is one that CCG was built to describe formally.

This convergence is, we believe, the more productive locus for understanding what LLMs do and what one can do with their output. Dependency grammar describes which words have been linked to which; CCG describes, in addition, \emph{what is still missing}, with what type, and \emph{how meaning will compose} when it arrives. Where dependency analysis recovers local structure, CCG recovers the predictive scaffolding of a compositional interpretation. For an autoregressive system whose entire operational character is predictive, this is not a marginal difference.

We exploit this convergence in a specific way. Rather than claiming that LLMs internally implement CCG --- a strong empirical claim we do not undertake to defend --- we treat CCG as a \emph{lifting} mechanism applied to LLM outputs. The idea is post hoc but principled. A sentence produced by an LLM is parsed into a CCG derivation by an existing CCG semantic parser of the kind exemplified by \texttt{ccg2lambda} \citep{martinez-gomez2016,mineshima2015}; the derivation yields a lambda-style logical form. The result is a pipeline that maps fluent-generated text into typed, compositional symbolic objects whose internal structure is transparent and auditable.

This framing carries three commitments worth stating up front. First, the framework is \emph{layered and modest}: dependency grammar gives a lightweight description of observable structure, and CCG gives a reconstructive layer that supports compositional interpretation. We do not claim that any one layer suffices, nor that the LLM ``contains'' any of them. Second, the framework is \emph{conservative}: nothing in it requires modifying the LLM, retraining it, or accessing its internals. The lifting operates on observable outputs. Third, the framework is \emph{coherent under alternative derivations}: a property we will return to in \cref{sec:bg-ccg}, because it is what makes the lifting well-defined in the presence of CCG's well-known proliferation of semantically equivalent surface analyses.

Although the paper enters through a debate about natural-language grammar, the framework it proposes does not stay there. Once the lifting is in place, two things follow. The operation it performs --- assigning a typed compositional structure to an incrementally produced expression --- is, through the Curry--Howard correspondence, the same operation as type-checking in programming languages; the framework therefore extends to the formal languages an LLM also produces, with the type system varying and the architecture held fixed. And the lifted structure, once available, supports checking at two levels: a compositional level, internal to the structure, and a content level that refers the structure outward to an external source of knowledge. These two developments --- the cross-domain reach and the two layers of checking --- are the substance of the paper's middle sections, and together they de-anchor the account from the cognitive assumptions that the dependency-grammar reply inherits. We take them up after first setting out the framework in its natural-language form.

It will help the reader to know in advance which questions the paper asks and how far it takes each of them, because the answers are of different kinds. We organize the paper around three research questions, and we state with each the sort of answer it receives: one is answered by argument, one by measurement, and one is opened rather than closed.

\begin{enumerate}[label=\textbf{RQ\arabic*.},leftmargin=*,itemsep=0.4em]
\item \emph{Is a typed compositional lifting of autoregressive LLM output well-defined?} That is: does the prefix-driven dynamic of autoregressive generation align with CCG's incremental, type-completing processing closely enough that a post hoc CCG lifting yields a typed semantic object which is coherent across alternative derivations and auditable down to its lexical and combinatory sources? This question is answered affirmatively, by argument, in \cref{sec:bg-ccg,sec:framework}. The argument rests on derivational equivalence modulo composition --- a formal property of the grammar --- and requires no claim about the model's internals.

\item \emph{Does the lifting behave as claimed on text that LLMs actually produce?} Coherence is a property of the grammar; whether an existing wide-coverage CCG parser and its semantic templates deliver it, at what coverage, and with what variation across the registers that LLM output spans, is an empirical matter. \Cref{sec:feasibility} reports a feasibility study on 2,922 sentences generated by four LLMs in five registers, organized around three hypotheses derived from the framework: (H1) two derivations of a sentence that differ only in their use of composition and type-raising receive the same lambda term; (H2) the lifted structure contains checkable --- saturated --- predications before the sentence is complete; (H3) lift coverage depends on register, not on the identity of the generating model. The study is a feasibility study, not a benchmark: it establishes that the substrate the framework requires exists and is usable, and it locates the instrument-side risks for the two directions that follow.

\item \emph{What does the lifted structure add, once available?} Two sub-questions fall here, and the paper opens both without closing either. First, what does the lifted structure contribute to checking generated content against external knowledge --- beyond what compositional analysis alone can do, and beyond what a mere superposition of a parser and a knowledge base would deliver? \Cref{sec:foundations} answers this at the architectural level, by the distinction between compositional and content checking and by the argument that incrementality yields the earliest structurally meaningful point of flagging, and illustrates it with worked examples; a quantitative evaluation against existing hallucination-detection methods is deferred, and its requirements are stated. Second, does the same operation extend to the formal languages an LLM also produces? \Cref{sec:beyond} argues that it does, on the warrant of the Curry--Howard correspondence, and sizes the work a first instantiation would take; no instantiation is built here.
\end{enumerate}

The distinction between these three kinds of answer is meant to be kept in view throughout. Where the text moves from what has been established to what has been argued for, and from what has been argued for to what has been opened, we say so; \cref{sec:discussion} returns to the three questions and states, for each, what the paper has and has not delivered.

The remainder of the paper is organized as follows. \Cref{sec:background} reviews the relevant background on dependency grammar, CCG, CCG-based semantic parsing, and neurosymbolic AI. \Cref{sec:framework} presents the two-layer lifting framework in its natural-language form, and \cref{sec:feasibility} reports a feasibility study of the lifting on LLM-generated text. \Cref{sec:foundations} sets out the framework's foundational commitments: the architectural rather than cognitive character of the stance, the distinction between compositional and content checking, and the role of external knowledge sources. \Cref{sec:beyond} argues that it does, on the warrant of the Curry--Howard correspondence; sizes the work a first instantiation would take; and reads a current dispute over the Lean formalization of an announced Navier--Stokes proof \citep{bastounis2026} as the compositional/content distinction operating in the wild, bounded by an impossibility result on faithful autoformalization that gives the deferral of this question theorem-grade warrant; no
instantiation is built here. \Cref{sec:synchronous} sketches a tighter, synchronous LLM--CCG coupling as one direction the framework opens. \Cref{sec:discussion} discusses what the framework claims and does not claim, its relation to existing work, and its limitations, and \cref{sec:conclusion} concludes.

\section{Background}
\label{sec:background}

\subsection{Dependency grammar as a descriptive layer for LLM behavior}
Dependency grammar represents syntactic organization through asymmetric head--dependent relations between lexical items, without an intervening layer of phrasal nodes. In its modern form it is associated with Tesni\`ere and with subsequent work in computational and cognitive linguistics, and it has become the dominant representational scheme in cross-linguistic corpus annotation through Universal Dependencies \citep{demarneffe2021}. The cognitive arm of this tradition, developed by Gibson and collaborators, emphasizes that dependencies impose a memory cost that scales with their length, yielding the empirical principle of \emph{dependency locality}: across languages and across constructions, word orders that minimize dependency length are preferred \citep{gibson1998,futrell2020}.

This framework has proved attractive as a descriptive lens on LLM behavior. The left-to-right realization of dependencies aligns naturally with autoregressive generation; many local syntactic regularities exhibited by LLMs can be characterized as the construction of head--dependent links one word at a time; and dependency-locality effects appear to be reproduced in the surprisal profiles of language models. We take this body of work as a well-motivated first-order account of \emph{what LLMs visibly do}. We do not contest it. We will argue, however, that it is not the deepest account available because it captures relations \emph{already realized} in surface text without explicit machinery for the predictive completion that drives generation.

\subsection{Combinatory Categorial Grammar}
\label{sec:bg-ccg}
Combinatory Categorial Grammar \citep{steedman2000,steedman2011} occupies a distinctive position among lexicalized grammatical formalizms. Its core intuition is that syntactic categories \emph{are} semantic types: every lexical item carries both a syntactic category and an associated lambda term, and a small set of combinatory rules --- forward and backward application, forward and backward composition, and type-raising --- determines both how categories combine and how meanings combine. Syntactic derivation and semantic composition proceed in lockstep. The classical warrant for this lockstep is Montague's algebraic conception of grammar, on which meaning assignment is a homomorphism from a syntactic algebra to a semantic algebra \citep{montague1970ug}; CCG can be seen as realizing that homomorphism in an incremental, surface-compositional form.

For present purposes, the crucial feature of CCG is its treatment of partial expressions. A sequence such as ``John loves'' is not a fragment without grammatical status; it has the category $\catS/\catNP$ (a sentence missing a noun phrase to its right) together with a partial lambda term $\lambda x.\,\textit{love}(\textit{john}, x)$ awaiting an argument. This is not a footnote to the formalizm: it is the formalizm's most distinctive commitment. Type-raising and composition were harnessed by Steedman and colleagues precisely to give a principled account of incremental, left-to-right interpretation, in which every prefix of a sentence carries a well-defined type and a well-defined partial meaning \citep{ades-steedman-1982,steedman2000}.\footnote{Type-raising itself has a dual ancestry worth recording. As a syntactic rule it is derivable as a theorem of the Lambek calculus \citep{lambek1958}; as a semantic operation it entered the field with Montague's lifting of individuals to generalized quantifiers, which renders proper names and quantified noun phrases combinatorially uniform \citep{montague1973}. CCG's distinctive contribution was to put the operation, together with composition, in the service of incremental derivation --- the point at which the logical and the semantic lineages converge.} In this respect CCG is unusually well aligned with the operational character of autoregressive language models, whose entire computation is predictive: at every position they implicitly carry a representation of how the prefix-so-far might be completed.

One feature of the formalizm is worth highlighting, as it recurs. The combinatory rules of CCG admit multiple derivation orders for the same string: a sentence such as ``John loves Mary'' can be derived either canonically, by first combining the verb with its object and then with its subject, or non-canonically, by type-raising the subject and composing it with the verb before applying to the object. The resulting surface derivations are distinct, but the lambda terms they produce coincide: composition is associative, and the two derivations yield equivalent terms in the underlying combinatory logic. This property, sometimes called \emph{derivational equivalence modulo composition}, was studied in detail in early work on efficient parsing of CCG, where it underwrites the design of chart-based parsers that can deliver all semantically distinct analyses without enumerating spurious ones \citep{pareschi-steedman-1987}. The same line has been pursued more recently in the design of fully incremental CCG parsers using tree-rotation operations \citep{stanojevic-steedman-2019,stanojevic-steedman-2020}, which maintain a fully connected, semantically interpretable structure at every word and which, in current implementations, use a neural supertagger as a front end. The supertagger's role deserves accurate statement, because it is easy to understate: beyond disambiguating among the known categories of seen words, modern supertaggers reliably predict categories for \emph{unseen} words from context, and the corresponding semantic category can then be supplied by rule. This generalization to unseen vocabulary is a powerful effect --- demonstrated when wide-coverage CCG parsing was ported to biomedical text \citep{rimell-clark-2009} --- and it is a substantial part of what makes current CCG parsers usable for practical semantic tasks, such as the construction of entailment graphs, on input far from their training distribution. We note these properties here because they are the formal facts that make a CCG lifting of an LLM prefix \emph{well defined}: alternative incremental derivations of the same prefix converge on the same typed semantic object, so the lifting does not inherit the prefix's derivational arbitrariness; and existing incremental parsers demonstrate that the symbolic substrate required to support such a lifting is implementable in practice.

\subsection{CCG-based semantic parsing}
\label{sec:bg-ccg-parsing}
CCG has served as the substrate for a substantial body of work on mapping natural language to logical form. Bos and collaborators developed the Boxer system, which combines a wide-coverage CCG parser with semantics in the style of Discourse Representation Theory \citep{bos2008}. Zettlemoyer and Collins introduced statistical CCG-based semantic parsers learned from logical-form annotations, thereby opening a long line of subsequent work on data-driven semantic parsing \citep{zettlemoyer2005}. More recent CCG parsers, such as EasyCCG and depCCG, provide fast and accurate wide-coverage analyses that are suitable as front-ends to logical interpretation \citep{lewis2014,yoshikawa2017}. The system most directly relevant to our pipeline is \texttt{ccg2lambda}, which composes lambda terms over CCG derivations using a template-based mapping from categories to semantic combinators, producing higher-order logical forms suitable for inference \citep{martinez-gomez2016,mineshima2015}.

These systems share a presupposition: they operate on \emph{curated sentences as input}, and their output is a logical form, typically used for natural-language inference, textual entailment, or knowledge-base population. We adopt their machinery but reposition it. The input we care about is \emph{LLM-generated} text, and the question is what is gained by treating the output of such a parser not as a final product but as a typed compositional representation of a generator's output --- a representation aligned, in its incremental structure, with the way the text was produced in the first place.

A parallel tradition in categorial semantics has pursued a quite different downstream object. Beginning with the distributional compositional categorical (DisCoCat) framework, and continuing through related work on the Lambek and Lambek--Grishin calculi, this line uses the categorial-grammar derivation as a recipe for composing distributional vector representations of word meaning into representations of phrase and sentence meaning, with grammatical types determining the tensor structure of the composition \citep{wijnholds-sadrzadeh-clark-2020}. The contrast is clarifying. DisCoCat-style approaches use the categorial skeleton to combine learned distributional representations and remain in vector space; the framework proposed here uses the same skeleton to lift LLM-generated text into symbolic lambda terms and exit vector space at the parser's output. Both routes exploit the same compositional logic, but to opposite ends. A long-standing concern for the vector-space route is the combinatorial blow-up of higher-order tensors required to type-match elements such as relative pronouns or quantifiers; this difficulty does not arise on the symbolic route taken here, which is one of its attractions in the present setting.

\subsection{CCG, neural language models, and incremental processing}
\label{sec:bg-ccg-llm}
The combination of CCG and large language models has recently been pursued in cognitive neuroscience, where the incremental parser of \citet{stanojevic-steedman-2019} has been combined with Transformer-based next-word predictability estimates to model neural signals during human sentence comprehension \citep{stanojevic-et-al-2023}. The aim of that work is different from ours --- to evaluate which grammatical formalizm better predicts brain responses to naturalistic speech --- but it is a useful demonstration that incremental CCG parsers can be productively coupled with language-model machinery, and that the resulting structure-building measures contain information not captured by the language model alone. We see the present paper as taking a complementary direction: rather than using CCG-derived measures alongside an LLM to model an external phenomenon, we use a CCG parser as a symbolic lifting layer applied to the LLM's own outputs, with the aim of producing typed compositional representations of generated content.

\subsection{Neurosymbolic AI and representational lifting}
Neurosymbolic AI is often presented as a broad family of approaches that integrate the flexibility of neural systems with the interpretability and inferential control of symbolic models. In practice, such integration can occur at very different levels of intimacy. Some approaches hybridize model internals, embedding symbolic constraints into training objectives or into the forward pass; others maintain a strict separation of layers, with a neural component producing material that a symbolic component then processes. The framework proposed here belongs to the second family. We do not modify the LLM. We treat it as a probabilistic generator whose outputs admit a grammatically informed symbolic reconstruction. In this sense, the CCG layer is best understood as a \emph{lifting} mechanism, as familiar in program analysis: a mapping from a less-structured representation to a more-structured one, preserving the information relevant to subsequent symbolic operations.

\section{A Two-Layer Lifting Framework}
\label{sec:framework}

We propose interpreting LLM outputs using two cooperating layers of analysis, applied post hoc to generated text.

\begin{enumerate}[leftmargin=2.2em]
    \item \textbf{Neural generation.} An LLM produces a token sequence $w_{1:n}$ probabilistically, conditioned on a prompt and on its learned distribution. We make no commitment about what occurs inside the model; we observe only its outputs.
    \item \textbf{Compositional CCG layer.} A CCG semantic parser assigns lexical categories and combinators to the generated text, yielding a derivation and an associated lambda term. This layer answers two questions that surface text does not: \emph{how does meaning compose}, and \emph{what is the type of what has been combined?}
\end{enumerate}

\subsection{Example: Incremental Lifting in Lockstep with Generation}
\label{sec:example}

To make the convergence between autoregressive generation and CCG's predictive dynamics concrete, consider an LLM emitting the tokens \textit{John}, \textit{loves}, \textit{Mary} one at a time. The example is small by design: \textit{John loves Mary} is the canonical CCG derivation that every reader knows, which lets attention fall on what is novel here --- the \emph{temporal} alignment between two systems running in lockstep.

At each generation step, two things happen in parallel. The LLM emits a token and updates its conditioning prefix; its next-token distribution then favors continuations compatible with that prefix. Independently, the CCG layer, taking the same prefix as input, resolves it to a typed category with a partial lambda term --- an object that makes explicit what kind of completion the prefix \emph{grammatically} requires. The two systems are not communicating, but their forward expectations agree.\footnote{A granularity caveat applies throughout. LLMs condition on subword tokens, while the CCG layer assigns categories to words. In the example each word is a single token, so the two systems align step by step; in general the alignment holds at word boundaries, with the lifting updating once per completed word rather than once per token. The distinction is immaterial for post hoc lifting, but it becomes operational in the synchronous regime of \cref{sec:synchronous}, which requires a buffering layer between token-level decoding and word-level typed state.}

\begin{center}
\renewcommand{\arraystretch}{1.4}
\small
\begin{tabular}{@{}clll@{}}
\toprule
\textbf{Step} & \textbf{LLM emits} & \textbf{LLM next-token expectation} & \textbf{CCG state for prefix} \\
\midrule
$t{=}1$ & \textit{John} & a verbal continuation is highly likely & $\catNP$,\;\;$\textit{john}$ \\[0.2em]
$t{=}2$ & \textit{loves} & a noun-phrase continuation is highly likely & $\catS/\catNP$,\;\;$\lambda y.\,\textit{love}(\textit{john}, y)$ \\[0.2em]
$t{=}3$ & \textit{Mary} & sentence is complete; punctuation expected & $\catS$,\;\;$\textit{love}(\textit{john}, \textit{mary})$ \\
\bottomrule
\end{tabular}
\end{center}

The crucial step is $t{=}2$. Once \textit{loves} has been emitted, the LLM is conditioning on the prefix \textit{John loves} and assigning high probability to noun-phrase continuations. At the same moment, the CCG layer has resolved the prefix to the type $\catS/\catNP$ with the partial term $\lambda y.\,\textit{love}(\textit{john}, y)$ --- a category that, by definition, denotes a sentence missing an $\catNP$ on the right. The LLM's distributional expectation and the CCG's typed expectation agree about what the prefix requires next, but for entirely different reasons: the LLM has learned the regularity statistically, while the CCG layer has derived it from lexical types and combinatory rules.

The CCG side of this alignment is built by type-raising (${>}\mathbf{T}$) and forward composition (${>}\mathbf{B}$), which together support a strictly left-to-right derivation that mirrors the order of emission. The static derivation is:

\vspace{0.5em}
\begin{center}
\renewcommand{\arraystretch}{1.2}
\begin{tabular}{ccc}
\textit{John} & \textit{loves} & \textit{Mary} \\
\cline{1-1}\cline{2-2}\cline{3-3}
$\catNP$ & $(\catS\backslash \catNP)/\catNP$ & $\catNP$ \\
$\textit{john}$ & $\lambda y\lambda x.\,\textit{love}(x,y)$ & $\textit{mary}$ \\
\multicolumn{1}{@{}c@{}}{\hrulefill\hspace{0.15em}${\scriptstyle{>}\mathbf{T}}$} & & \\
$\catS/(\catS\backslash \catNP)$ & & \\
$\lambda p.\, p(\textit{john})$ & & \\
\multicolumn{2}{@{}c@{}}{\hrulefill\hspace{0.15em}${\scriptstyle{>}\mathbf{B}}$} & \\
\multicolumn{2}{c}{$\catS/\catNP$} & \\
\multicolumn{2}{c}{$\lambda y.\,\textit{love}(\textit{john}, y)$} & \\
\multicolumn{3}{@{}c@{}}{\hrulefill\hspace{0.15em}${\scriptstyle{>}}$} \\
\multicolumn{3}{c}{$\catS$} \\
\multicolumn{3}{c}{$\textit{love}(\textit{john}, \textit{mary})$} \\
\end{tabular}
\end{center}
\vspace{0.5em}

Two observations follow. First, by the derivational-equivalence property recalled in \cref{sec:bg-ccg}, this left-to-right derivation yields the same final lambda term as any other CCG derivation of the same sentence, including the canonical bottom-up application-only derivation. The post-hoc lifting is well-defined regardless of which derivation order the parser internally follows. Second --- and this is the point that the parallel table is meant to bring out --- the prefix-by-prefix structure of the left-to-right derivation is not merely \emph{available} in CCG; it is \emph{aligned with} the temporal order in which the LLM produced the text. The lifting is not only consistent with generation; it is \emph{operationally aligned} with it, recovering at each prefix exactly the typed object whose unsaturated argument matches what the LLM's distribution expects to come next.

A dependency analysis can be inserted as a descriptive intermediate to identify the local head--dependent organization of the generated string. We treat it here as informative scaffolding rather than as a separate computational layer of the pipeline; the substantive lifting is performed by the CCG analysis.

The architecture is intentionally modest in its ontological commitments. It does not claim that LLMs internally store symbolic grammatical rules, nor that the agreement between distributional and typed expectations exhibited in the table above reflects a shared mechanism --- only that the two expectations coincide at each prefix, for reasons that the formal alignment between CCG and prefix-driven processing explains. It does not require that every fluent output admit an unproblematic compositional reconstruction; sentences may be ambiguous, underspecified, or pragmatically dependent, and the CCG layer will fail or yield multiple analyses in such cases just as it does for human-authored text. What it does claim is that, for an interesting class of outputs --- in particular, declarative and argumentatively organized statements --- the lifting yields typed compositional representations that are stable, transparent, and available for downstream symbolic processing.

Two properties of the resulting representation deserve emphasis. First, the lifting is \emph{coherent}: by the derivational-equivalence property just recalled, the lambda term assigned to a sentence does not depend on which of the (potentially many) CCG derivations the parser follows. Second, the lifting is \emph{auditable}: every component of the resulting term traces back either to a lexical entry or to a CCG combinatory step, with no opaque inference between the surface string and its compositional reading.

A distinction should be drawn, however, between the ambiguity the coherence property disposes of and the ambiguity it does not. Derivational equivalence guarantees that \emph{spurious} ambiguity --- alternative derivation orders of the same analysis --- leaves the lifting well-defined. \emph{Genuine} ambiguity --- lexical category choice, attachment, quantifier scope --- is another matter: distinct analyses yield genuinely distinct lambda terms, and the lifting of an ambiguous sentence is a set of typed candidates rather than a single term. This is not a defect of the formalizm (the sentence \emph{is} ambiguous), but it obliges the architecture to state a policy, particularly for the checking layers developed in \cref{sec:foundations}: content checks should fire only on predications that are stable across the parser's $n$-best analyses, or on the supertagger-disambiguated analysis weighted by its confidence, with hard interventions --- in the synchronous regime of \cref{sec:synchronous} --- reserved for flags that survive this discipline. We record the policy here because it is what prevents a mid-sentence misattachment from producing a false flag with downstream consequences for generation.

\subsection{Why LLM-generated text, and not just any text}
\label{sec:why-llm}

A natural question is whether anything in the framework above is specific to LLM-generated text, or whether it is simply semantic parsing applied to a particular input distribution. The machinery is indeed the same as that used for human-authored text. What is different is the failure mode the input distribution exhibits, and what an auditable compositional lifting offers in response to it.

LLM-generated text exhibits a characteristic pathology: the confident production of content that is not entailed by anything the system has access to, including its own prior context. Human text is not free of confident unsupported assertion, but the failure differs in distribution and in detectability: in LLM output it is systematic rather than exceptional, arises across every register, and arrives unmarked by the hedges, sourcing behaviors, and stylistic tells that often accompany it in human prose. This pathology --- variously labelled hallucination, confabulation, or false-positive generation --- is the principal obstacle to the use of LLM output in settings where the cost of an unverified claim is high. Purely distributional methods for detecting it have so far made limited progress, in part because the relevant defect is not an aberrant surface feature but the absence of any verifiable compositional support for what is being said. The output is locally well-formed; it is in its global semantic dependencies that it fails to cohere.

This is the setting in which an auditable compositional lifting earns its keep. A typed lambda term derived from generated text exposes precisely the dependencies that hallucination obscures: which entities are bound, which predicates are applied to them, and which propositional commitments follow. Successive lifted representations, taken across a passage, can in principle be checked for compositional coherence, and --- as we develop in \cref{sec:foundations} --- the same lifted structure is what makes a further check against external knowledge sources possible at all. We do not develop a hallucination-detection method here; we note only that lifting LLM output into typed compositional form is one of the few moves that opens such a method to construction, and that the case for it is significantly stronger in the LLM-as-input setting than in the human-text setting where compositional parsing was originally developed.
\section{A Feasibility Study of the Lifting}
\label{sec:feasibility}

This section reports the measured answer to RQ2: whether the lifting of \cref{sec:framework} behaves as claimed on text that LLMs actually produce. It is a feasibility study of the substrate the framework requires, not an evaluation of the checking layers built on it, which belong to RQ3. The protocol --- corpus design, filters, metrics, and success criteria --- was fixed in the study repository before any text was generated (\cref{sec:feas-repro}). The audit of the first full run led to four documented amendments, all post hoc and all reported as such; where an amendment changed a metric, the original value is reported alongside the amended one.

\subsection{Setup}
\label{sec:feas-setup}

\paragraph{Corpus.} Four generating models, one current model per provider and one open-weights model (GPT-6.1 Sol, Claude Sonnet 5.5, Gemini 4 Carbon, Llama 4 Maverick), were each prompted in five registers ordered a priori by distance from the newswire register of the parser's training data: news (R1), explanatory (R2), argumentative (R3), technical-scientific (R4), and conversational (R5). A bank of 25 topics was crossed with models and registers, giving 14 generations per model--register cell and 280 generations in all. Sentence segmentation and pre-registered filters (declaratives of 5--40 tokens, no lists, code, or URLs; duplicates removed) kept 2,922 sentences and rejected 152. Decoding could not be held identical across providers: the reasoning-tier models were used at their default sampling, the others at temperature 0.7; the parameters actually sent are logged with every generation.

\paragraph{Instrument.} Sentences were parsed with depccg \citep{yoshikawa2017}, a wide-coverage A* CCG parser, requesting the 25 best derivations, and interpreted with the ccg2lambda semantic templates \citep{martinez-gomez2016} distributed with depccg. Two template sets were compared: the classic higher-order templates (emnlp2015) and the Neo-Davidsonian event templates, in which verbs introduce an event variable and their arguments enter through thematic roles, $\mathit{Subj}(e) = x$, $\mathit{Acc}(e) = y$ \citep{parsons1990}. We say a sentence is \emph{lifted} when the 1-best derivation receives a well-formed logical form. On 32 sentences (1.1\%), all rich in comma-separated coordination, the 25-best search exceeded its time budget; these were parsed at 1-best and excluded from analyses that need alternative derivations.

\paragraph{Derivation pairs for H1.} Testing H1 requires pairs of derivations that differ \emph{only} in their use of composition and type-raising. The parser's own n-best list does not supply them: depccg is a normal-form parser \citep{eisner1996}, built to avoid precisely the spurious derivations whose semantic equivalence H1 asserts, so its alternatives are genuinely different readings (we found, for instance, prepositional attachments differing under identical supertag sequences). We therefore constructed the pairs. Following the tree-rotation analysis of incremental CCG parsing \citep{stanojevic-steedman-2019}, and in the line of incremental chart parsing with type-raising and composition \citep{pareschi-steedman-1987}, each 1-best normal-form derivation was rewritten into an incremental one by two rules, applied exhaustively bottom-up:
\begin{align*}
&\text{(R1)}\quad [\,X/Y\;\; [\,Y/W\;\; W\,]_{>}\,]_{>} \;\Rightarrow\; [\,[\,X/Y\;\; Y/W\,]_{>\mathbf{B}}\;\; W\,]_{>}\\
&\text{(R2)}\quad [\,\mathit{NP}\;\; [\,(S\backslash\mathit{NP})/W\;\; \ldots\,]\,]_{<} \;\Rightarrow\; [\,\mathbf{T}(\mathit{NP}) = S/(S\backslash\mathit{NP})\;\; [\ldots]\,]_{>}
\end{align*}
R1 (and its variant with $W/Z$ for $W$ and composition at the top) composes left material before the material to its right is complete; R2 type-raises a subject so that R1 can compose it with its verb. R2 applies only when the verb phrase is built by forward application of $(S\backslash\mathit{NP})/W$ to its argument; a verb phrase in which the verb is itself the argument of a backward modifier $X\backslash((S\backslash\mathit{NP})/W)$ is left in place, since raising and composing there would not preserve the semantics. Both rules preserve lexical categories and predicate--argument dependencies, so a normal-form derivation and its rotation differ in derivation shape and nothing else; both are interpreted by the same templates. The rotation is partial: backward modifiers, coordination, and punctuation are left in place, so the rotated derivation is one incremental derivation licensed by the grammar, not the maximally incremental one.

\paragraph{Availability (H2).} An atomic predication $P(a_1,\ldots,a_k)$ in the term of a constituent is \emph{saturated} when none of its arguments is still bound by a lambda: every argument is a constant or a quantified variable. Saturation is the point at which a predication becomes a checkable claim. The availability of a predication of the final logical form is the earliest end position of a constituent in which it occurs saturated, and its \emph{lead} is the fraction of the sentence (trailing punctuation excluded) that remains after that point. We report properties (arity 1) and relations (arity $\ge 2$, thematic roles included) separately, and we control for a lexical effect: a predication saturated inside a single word's own entry (as event predicates are in the event templates) is available at that word's position by construction, not by composition.

\subsection{Coverage (H3)}
\label{sec:feas-coverage}

Parse coverage is uninformative for depccg, which returned a derivation for every sentence. Lift coverage is not (\cref{tab:feas-lift}). With the emnlp2015 templates only 40.3\% of sentences were lifted. A re-parse of failing sentences with error logging located the cause: these templates treat punctuation and compound markers as lexical constants ($\mathit{\_COMMA}$, $\mathit{\_HYPHEN}$, possessive \emph{'s}) and then apply them as predicates, yielding ill-formed terms on comma-coordinated modifiers, hyphenated compounds, and possessives, all frequent in LLM prose. With the event templates, applied to the same parse trees, 97.7\% of sentences were lifted. The event templates were adopted as the primary instrument (amendment A3); because syntax does not depend on the templates, this required re-interpreting the cached derivations, not re-parsing, and re-interpreting with the original templates reproduced the original lift rate exactly (0.403).

\begin{table}[t]
\centering
\caption{Lift coverage (share of sentences whose 1-best derivation receives a well-formed logical form), by register and by generating model, for the two template sets. Same parse trees throughout.}
\label{tab:feas-lift}
\begin{tabular}{@{}lrrr@{}}
\toprule
 & $n$ & emnlp2015 & event \\
\midrule
R1 news                 & 622 & 48.7\% & 98.6\% \\
R2 explanatory          & 662 & 42.7\% & 97.7\% \\
R3 argumentative        & 620 & 35.2\% & 97.4\% \\
R4 technical-scientific & 427 & 26.2\% & 94.6\% \\
R5 conversational       & 591 & 44.2\% & 99.5\% \\
\midrule
M1 GPT-6.1 Sol          & 915 & 40.4\% & 98.1\% \\
M2 Claude Sonnet 5.5    & 659 & 40.2\% & 97.9\% \\
M3 Gemini 4 Carbon      & 699 & 39.2\% & 97.1\% \\
M4 Llama 4 Maverick     & 649 & 41.3\% & 97.7\% \\
\midrule
All                     & 2,922 & 40.3\% & 97.7\% \\
\bottomrule
\end{tabular}
\end{table}

H3 predicted that coverage degrades with distance from the parser's training register and not with the identity of the generating model. We tested it with a binomial mixed model (lift $\sim$ register + model + sentence length, random intercept for topic; variational Bayes) and, as a sensitivity check, a logistic GLM with topic fixed effects and standard errors clustered by generation. With the event templates the second half of H3 holds: no model differs reliably from another (largest posterior mean $0.28$, posterior SD $0.26$). The first half does not: there is no gradient across the pre-registered register ordering. Only the technical-scientific register lifts less (posterior mean $-0.74$, SD $0.21$; GLM $p = 0.034$), conversational text lifts best, and sentence length has the expected negative effect ($-0.37$ per SD). With coverage this close to ceiling the test has little power, and we read it as showing that the residual failures concentrate in technical prose rather than as evidence about register distance in general. Under the emnlp2015 templates, by contrast, raw lift rates were nearly identical across models (39--41\%) but, controlling for length, the two models that write longer sentences lifted more ($+0.50$ and $+0.70$); with the event templates this conditional model effect disappears. An instrument with coverage gaps can thus produce apparent differences between generators that a more complete instrument does not show.

\subsection{Term invariance (H1)}
\label{sec:feas-invariance}

The rotation applied 29,049 rotations and 3,390 type-raisings, changing the derivation of 2,880 of the 2,922 sentences, with no category errors. Rotated derivations were lifted as often as normal-form ones (97.8\% against 97.7\%). Of the 2,814 rotated sentences lifted under both derivations, 2,790 (99.15\%) received strictly identical logical forms, identity being decided after $\beta$-reduction, $\eta$-reduction, and canonical renaming of bound variables. The remaining 24 are malformed under \emph{both} derivations (21 ill-typed, 3 unreadable): they are defects of the templates, 83\% of them on sentences with comma-separated coordination, which the rotation did not introduce. Among the pairs in which both logical forms are well-formed, invariance is complete (2,790 of 2,790), and it holds in every register and for every generating model.

H1 is thereby confirmed in the form the framework needs: when two derivations of the same sentence differ only in composition and type-raising, the existing instrument assigns them the same lambda term. The argument for RQ1 rested on derivational equivalence as a property of the grammar; the study shows that the property survives its implementation in a wide-coverage parser and a set of hand-written templates, on text no linguist wrote.

\subsection{Availability before completion (H2)}
\label{sec:feas-availability}

The registered form of H2 concerned the stabilization of prefix categories. Measured by re-parsing every prefix of a 500-sentence subsample, it proved uninformative: the parser must commit to a root category, and it coerced 77\% of prefixes into complete declarative sentences, so that every prefix appeared ``connected'' by construction. We therefore replaced the measure (amendment A2) by the availability of saturated predications defined in \cref{sec:feas-setup}, which asks the question H2 was meant to answer: how early in the sentence does the lifted structure contain checkable claims? \Cref{tab:feas-avail} reports it for the normal-form derivations and their rotations.

\begin{table}[t]
\centering
\caption{Availability of saturated predications before sentence end (event templates, 2,814 sentences lifted under both derivations). ``Before end'': share of predications saturated before the last word; ``lead'': mean fraction of the sentence remaining at that point.}
\label{tab:feas-avail}
\begin{tabular}{@{}lrrrrr@{}}
\toprule
 & & \multicolumn{2}{c}{normal-form} & \multicolumn{2}{c}{rotated} \\
\cmidrule(lr){3-4}\cmidrule(l){5-6}
 & $n$ & before end & lead & before end & lead \\
\midrule
Properties (arity 1)             & 37,893 & 80.8\% & 0.42 & 83.4\% & 0.43 \\
\quad of which composed          & 26,102 & 74.0\% & 0.38 & ---    & ---  \\
Relations (arity $\ge 2$)        & 9,280  & 54.9\% & 0.29 & 66.9\% & 0.36 \\
\midrule
Connected prefixes               & ---    & \multicolumn{2}{c}{15.2\%} & \multicolumn{2}{c}{23.5\%} \\
\bottomrule
\end{tabular}
\end{table}

Three results stand out. First, most of the content of a sentence becomes checkable before the sentence ends. Even discounting the 31\% of properties saturated within a single lexical entry, which are available at their word's position by construction, three quarters of the properties that composition builds are saturated before the last word, on average with 38\% of the sentence still to come. Second, relations are later than properties, and for a principled reason: a relation is saturated only when its last argument has been supplied, and in English main clauses the last argument tends to come late. This is a fact about the grammar of the claims, not about any derivation. Third, both the semantic representation and the derivation move this boundary. Neo-Davidsonian decomposition splits a relation into thematic roles that saturate separately, $\mathit{Acc}(e,y)$ as soon as the verb phrase is complete, before the subject relation: relations saturated before sentence end rise from 24.2\% under the emnlp2015 templates to 54.9\% under the event templates (the former measured on the 40\% of sentences those templates lift, so the comparison is indicative). The incremental derivation then raises the figure to 66.9\%: rotation makes 24.3\% of relations available earlier and none later, with the largest gain in conversational text (57.8\% to 74.3\%) and the smallest in technical text (56.7\% to 64.3\%).

The earliest-flagging argument of \cref{sec:foundations-incrementality} is therefore borne out with a precision the argument itself did not have. The structurally earliest point at which a claim can be checked is the point at which it is saturated; for properties of entities this is typically the close of the noun phrase, early in the sentence, which is where the content failure of the Zeus example sits (\cref{sec:example}); for relations it is the arrival of the last argument, and how early that is depends on the semantic representation and on the incrementality of the derivation, both of which the lifting controls.

\subsection{What the study does and does not show}
\label{sec:feas-limits}

The study establishes that the substrate the framework requires exists and is usable on LLM output: a wide-coverage parser covers it, a suitable template set interprets nearly all of it, derivational equivalence holds in the implementation, and checkable predications are available well before sentence completion. It does not establish anything about the checking layers built on that substrate, which remain the subject of RQ3. Several limitations qualify the results. All four amendments were made after the full run and in response to its audit; the original metrics are retained in the repository with the amended ones. The rotation is partial and exercises one incremental derivation per sentence: punctuation is not reattached to its left neighbor, which would remove many of the breaks in left-branching that commas now create, and type-raising is restricted to $S$ results, whose relaxation would give more incrementality still. H1 is tested on the logical form of the whole sentence; comparing the meaning of every constituent that survives the rotation would be a stronger test, and is the natural next step. All results come from a single parser and from one family of semantic templates, whose remaining defects (coordination and generalized quantifiers such as \emph{most}) are documented above. The 1-best derivation is taken as given: we measure the lifting of the parser's analysis, not the correctness of that analysis. Decoding settings differed across providers. The corpus is English, and its five registers are prompted, not naturally occurring. Finally, three engineering findings are worth recording for anyone repeating the study: the 25-best search explodes on sentences with long coordinated lists; depccg's parallel parsing path can deadlock once PyTorch is loaded, so batches must be kept small; and the coverage of the semantic templates, not of the parser, is the binding constraint on lifting LLM prose.

\subsection{Reproducibility}
\label{sec:feas-repro}

Code, prompts, the frozen corpus (with a checksum recorded before analysis), all parses, per-sentence results, the protocol, and the four amendments are available in the study repository (see the Data Availability statement). The pipeline runs in a Docker image on a CPU-only machine; the full 25-best parse of the corpus took about 19 hours, and every subsequent analysis, including the template switch and the rotation experiment, runs on the cached parses in minutes. A maintained implementation of tree rotation for CCG derivations is available in the \texttt{ccgtools} library (\url{https://github.com/stanojevic/ccgtools}); our rotation is an independent reimplementation of the two rules stated above.

\section{Foundations}
\label{sec:foundations}

The framework, as developed so far, takes natural language as input and CCG as a typed compositional standard. But nothing prevents us from going further and fully embracing the generative capabilities of LLMs. The cross-domain extension we develop in the next section --- including programming languages, description logics, query languages, and even the languages of formal mathematics --- raises the question of which theory of language is at play. The dependency-grammar perspective we used as a starting point in \cref{sec:intro} is located within cognitive linguistics and, while elegant and compelling, is anchored in human processing: it is imbued with the assumption that dependency-locality effects exist because human memory imposes costs on long-range dependencies, and that the empirical regularities follow from this cognitive constraint. The framework we develop here, however, will apply equally to programming languages --- as in the Solidity case exemplified below --- and to description logics such as OWL: in both cases, artefacts whose ``processing'' has nothing to do with human memory. Paradoxical as it may seem at first glance, leveraging the methodological approach proposed thus far to its full capabilities implies a lightening, rather than a burdening, of the foundational assumptions on which it rests.
 
\subsection{An architectural stance, not a cognitive one}
\label{sec:foundations-architectural}
 
The framework's central commitment is to a specific architectural pattern: a probabilistic prefix-driven generator whose output is checked, prefix by prefix, by a typed compositional analyzer.\footnote{We use \emph{prefix-driven} rather than \emph{left-to-right} as the framework's primary technical vocabulary, because the architectural commitment is to the conditioning relation by which each successive output is determined by the prefix so far, not to any particular writing direction. Left-to-right is the writing-system-specific instance for English, Solidity, and the other languages we predominantly engage with; right-to-left, vertical, and non-sequential writing systems realize the same prefix-driven structure differently. The framework is indifferent to the realization and depends only on the structure.} The privileged producer in this pattern is the large language model. The framework exists because LLMs generate text token by token, conditioned on the prefix produced so far, and because this mode of generation is both powerful and fallible in characteristic ways. Compositional incoherence, content drift, and the particular kind of confident-sounding error we call hallucination are all symptoms of probabilistic prefix-driven production. The framework's claim is that CCG-style typed lifting is architecturally well suited to checking the output of such a producer, because it operates at the same granularity --- the prefix --- and in the same direction as the production.
 
The cross-domain reach of the framework, developed in \cref{sec:beyond}, is not about parallel producers but about parallel \emph{output languages}. The same LLM can produce English, Solidity, OWL, or SQL; in each case it produces prefix by prefix; in each case its output is subject to the same kind of probabilistic compositional failure. What differs across these cases is the type system against which the output is to be checked, and the body of external knowledge that informs the content layer of the check. The producer is held constant; the output language varies. This is the genuinely interesting locus of the framework's cross-domain claim, and it is sharper than a claim about parallel producers would be. A deterministic Solidity compiler does not need this kind of checking architecture, because its output is by construction type-correct; an LLM producing Solidity does need it, because its output is type-correct only probabilistically. The framework is about LLMs as producers across output domains, not about any typed compositional producer whatsoever.
 
This re-anchoring also clarifies the relationship to the cognitive-linguistic perspective that opened the paper. The Gibsonian story is an account of \emph{human} prefix-driven production, and it has been our entry point precisely because human language production exhibits the same prefix-by-prefix character that LLMs exhibit by other means. The framework here generalizes not by abstracting away from the LLM but by holding the LLM as the producer and varying the output language. Gibson's account survives as a complementary cognitive-linguistic description of one specific producer --- the human speaker --- whose generative profile is structurally similar to the LLM's. The framework does not depend on this similarity being more than structural, and it does not depend on any further theory of human cognition.\footnote{This methodological posture has a recognizable precedent in Dennett's \emph{intentional stance} \citep{dennett1987}, on which a system may be characterized by the stance that is analytically useful toward it --- there, treating it as having beliefs and desires for predictive purposes --- without commitment to claims about its intrinsic nature. The stance we adopt here is architectural rather than intentional, but the licensing move is the same: a producer is characterized by its generative profile, not by claims about whether it has a mind.} What it does depend on is the prefix-driven character of the producer.

The stance has a deliberately cooling consequence for a debate that currently runs hot. Much discussion treats LLM linguistic fluency as a proxy battlefield for the question of mind --- either as evidence that these systems understand, or as an occasion to insist that, lacking minds, they cannot genuinely be doing language at all. The framework declines both inferences, and in doing so recovers two older postures at once. It recovers Montague's anti-psychologism --- the insistence that the theory of language, natural or formal, is a branch of mathematics rather than of psychology \citep{montague1970ug} --- and extends it from the theory of meaning to the characterization of producers: what the framework requires of a producer, human or artificial, is a generative profile, not a mind. And it echoes Turing's deflationary gesture of replacing a metaphysical question he judged ``too meaningless to deserve discussion'' with an operational one \citep{turing1950} --- while going a step further in one respect: where Turing retained linguistic behavior as the criterial proxy for intelligence, the present account removes language production from that battlefield altogether. On this view, fluent production neither requires a mind nor evidences one; it requires a prefix-driven generative profile, which humans and LLMs realize by different means.
 
This last point deserves to be stated positively rather than left as a silent assumption. The framework's architectural choice is not universal across all conceivable producers. A producer that emits language in a non-sequential mode --- as a complete graph, as a one-shot semantic structure, as a globally optimized whole --- would not be an instance of the pattern, and the framework would not apply to its output in the form developed here. The framework is bounded to producers whose generative profile is prefix-driven, and within that bound, it makes a precise architectural claim. Current LLMs fit the bound; future systems that depart from prefix-driven generation would require a different checking architecture. The framework's reach is therefore historically situated rather than metaphysically grounded: it is a claim about a specific architectural fit between a class of producers we presently have and a class of compositional analyzers that match their generative profile.

Because the architectural stance disclaims commitments about model internals, it is worth noting that recent interpretability evidence supports the abstention rather than merely tolerating it. \citet{gurnee2026} present causal evidence that LLMs maintain a small, privileged set of internal representations --- a ``global workspace'' of word-linked, verbalizable patterns --- that mediates deliberate multi-step reasoning: intermediate concepts appearing in neither prompt nor output can be identified in this space and swapped, with the swap propagating coherently to the several downstream computations that read from the same representation. Two aspects of these findings bear on the present framework, in opposite directions. On one side, ablating the workspace leaves fluent generation and shallow classification largely intact while collapsing multi-step reasoning --- a mechanistic dissociation between surface fluency and compositional coherence that is precisely the premise of \cref{sec:why-llm}: the characteristic pathology of confident generation is not an aberrant surface feature but a failure at the level of semantic dependencies that fluency alone does not carry. On the other side, the representations in question are atomic and untyped --- word-linked patterns without categories, without lambda terms, without combinatory structure --- so nothing in the findings suggests that the model internally realizes the typed composition our lifting recovers. We draw the moral accordingly: on current evidence, the internal workspace holds word-like atoms, and the compositional algebra over those atoms is exactly what an external typed lifting supplies. The evidence thus locates the framework's contribution rather than licensing a stronger claim, and it gives empirical texture to a division of labor that the architectural stance had adopted on principle.
 
\subsection{Production and knowledge of appropriate use}
\label{sec:foundations-production-and-knowledge}
 
The architectural pattern has two components. There is the \emph{production} of expressions, performed by the LLM in the role established above, and there is \emph{knowledge of appropriate use} against which the produced expressions can be checked. The production component is held constant across our cases of interest: it is the LLM, generating prefix by prefix, fallibly, and across whichever output language it has been prompted to produce. What varies is the body of knowledge against which the output is checked. When the LLM produces natural language, the relevant knowledge of appropriate use comprises grammatical, semantic, and pragmatic constraints, together with whatever world-knowledge is needed to evaluate referential content. When it produces a program in a typed language, the relevant knowledge comprises the language's type system and execution semantics, together with the body of practical and security knowledge that experts in that language maintain. When it produces an expression in a description logic, the relevant knowledge comprises the ontology's formal axioms together with the external content --- lexical, taxonomic, factual --- that fixes what the concepts in the ontology actually denote.
 
This bipartite structure is not a coincidence. Any language whose expressions are meant to bear a determinate relationship to a domain --- whether the domain is the world, the state of a blockchain, or the structure of a knowledge graph --- requires both the production of expressions and the standard against which their appropriateness is judged. The novelty of the present framework is not the recognition of this bipartite structure, which is ancient, but the specific computational realization it proposes: production is a probabilistic neural process operating prefix by prefix, and knowledge of appropriate use is implemented by a typed compositional analyzer running on the produced output --- post hoc in the conservative version, in parallel with generation in the synchronous variant of \cref{sec:synchronous}.
 
\subsection{Two layers of checking: compositional and content}
\label{sec:foundations-two-layers}
 
Knowledge of appropriate use, in this framework, has two layers, and being clear about the distinction between them is essential to avoiding overclaim. The typed compositional lifting catches \emph{compositional} failures: cases where the parts of an expression do not combine into a coherent whole, where required arguments are missing, where the types of subexpressions do not match the types their context demands. This is the kind of checking that proceeds entirely from the structure of the expression and the formal type system it is judged against. The Solidity example we develop in the next section illustrates this layer: a function that omits a required \texttt{require} statement produces a lifted type whose effect component is strictly weaker than the contract demands, and the discrepancy is visible at the level of type-comparison alone.
 
The lifting does not, however, catch \emph{content} failures: cases where an expression is compositionally well-formed but factually wrong about the domain. The sentence \emph{``The Aztec god Zeus loved Europa''} is a case in point: a CCG derivation proceeds without difficulty, and the resulting lambda term is well-typed, yet the sentence is wrong, and wrong in a way that a structurally identical lifting of \emph{``The Greek god Zeus loved Europa''} is not. The wrongness lives in the lexicon --- in the fact that Zeus belongs to the Greek pantheon, not the Aztec one --- which is to say, in an external knowledge source that the typed lifting alone cannot access. A CCG parser equipped only with syntactic categories and types of standard generality cannot make this distinction: current wide-coverage CCG parsers are limited in their access to lexical semantics, and supertaggers trained on category disambiguation do not by themselves carry the kind of lexical-semantic content that would resolve cases of this type (M. Steedman, personal communication, May 2026). We work this example through in detail at the end of the subsection.
 
The same distinction appears in formal languages. A Solidity function --- Solidity being the principal language for the programs, known as smart contracts, that run on blockchains such as Ethereum --- may type-check cleanly while exhibiting a \emph{reentrancy} pattern: it makes an external call before updating its own internal state, so that the called party can call back into the original function and act on state that has not yet been updated. This is a famous failure mode, and the one behind the 2016 collapse of ``The DAO,'' an Ethereum investment fund whose contract was drained of roughly a third of its holdings through exactly this mechanism. Crucially, reentrancy is not visible at the level of types alone: the function compiles, the values flow through the type system without contradiction, and the lifted lambda term is well-formed. The failure is about the \emph{ordering} of effects, against a body of \emph{security knowledge} about Solidity that is external to its type system. Similarly, an OWL class expression asserting that cows are carnivores is syntactically well-formed and may pass description-logic consistency checking depending on the ontology in scope, but is wrong against any plausible biological ontology --- against external knowledge of what cows actually are.
 
In each case, the typed compositional lifting does not directly catch the failure. What it does is more modest and more interesting: it produces a structured object --- a typed lambda term, a categorial derivation, an explicit representation of compositional dependencies --- against which a content check, performed against an external knowledge source, becomes operationally available. The lifting does not perform the check; it produces the substrate on which the check can be formulated.

We now work through the Zeus example, both to see the two layers operating together and to see precisely where the CCG layer earns its place. The two sentences \emph{``The Aztec god Zeus loved Europa''} and \emph{``The Greek god Zeus loved Europa''} differ in a single modifier. The compositional lifting proceeds identically for both, prefix by prefix. We show the incremental categories and partial logical forms; the combinatory steps are as in \cref{sec:example} and are elided for compactness. Two lexical entries deserve explicit statement. The modifier \textit{Aztec} carries the standard adnominal category $\catN/\catN$, with term $\lambda P\lambda x.\,\mathrm{aztec}(x)\wedge P(x)$; its forward composition with the determiner's $\mathit{NP}/N$ yields the $\mathit{NP}/N$ prefix category of the second row. And the fourth row treats the bare proper name as an identificational nominal modifier --- an entry of category $\mathit{NP}\backslash \mathit{NP}$ contributing the conjunct $x = \textit{zeus}$ --- which selects the restrictive reading of the construction; an appositive reading would require a different entry. Nothing downstream depends on either choice, but both should be visible.

\begin{center}
\renewcommand{\arraystretch}{1.3}
\small
\begin{tabular}{@{}r l l@{}}
\toprule
\textbf{prefix} & \textbf{category} & \textbf{partial logical form} \\
\midrule
\emph{The} & $\mathit{NP}/N$ & $\lambda P.\,\iota x.\,P(x)$ \\
\emph{The Aztec} & $\mathit{NP}/N$ & $\lambda P.\,\iota x.\,\mathrm{aztec}(x)\wedge P(x)$ \\
\emph{The Aztec god} & $\mathit{NP}$ & $\iota x.\,\mathrm{aztec}(x)\wedge \mathrm{god}(x)$ \\
\emph{The Aztec god Zeus} & $\mathit{NP}$ & $\iota x.\,\mathrm{aztec}(x)\wedge \mathrm{god}(x)\wedge x=\mathrm{zeus}$ \\
\midrule
\emph{\ldots loved} & $(S\backslash \mathit{NP})/\mathit{NP}$ & $\lambda y\lambda x.\,\mathrm{love}(x,y)$ \\
\emph{\ldots loved Europa} & $S\backslash \mathit{NP}$ & $\lambda x.\,\mathrm{love}(x,\mathrm{europa})$ \\
\emph{(complete)} & $S$ & $\mathrm{love}(\iota x.[\mathrm{aztec}(x)\wedge\mathrm{god}(x)\wedge x{=}\mathrm{zeus}],\,\mathrm{europa})$ \\
\bottomrule
\end{tabular}
\end{center}

The type-check succeeds for both sentences: each is compositionally well-formed, and the only difference between the two derivations is the substitution of $\mathrm{greek}$ for $\mathrm{aztec}$ in one conjunct. Compositional checking cannot distinguish them; that is the point. The content check operates on the structure the lifting has exposed. At the fourth row --- the prefix \emph{The Aztec god Zeus}, \emph{before the verb has been read} --- the partial logical form already contains the conjunction $\mathrm{aztec}(\mathrm{zeus})\wedge\mathrm{god}(\mathrm{zeus})$. This conjunction is a queryable proposition. Mapping it to a query against a lexical ontology --- itself a step a full implementation must specify, not one the lifting performs automatically --- yields the lookup $\textsc{pantheon}(\mathrm{zeus})\overset{?}{=}\textsc{aztec}$, against which the ontology returns $\textsc{pantheon}(\mathrm{zeus})=\textsc{greek}$. The asserted modifier and the retrieved fact are inconsistent, and the content layer raises a flag. For the second sentence, the same lookup returns a match and the prefix is cleared. The flag is raised by the ontology, not by the type-checker: the lifting produced the queryable atom; the external source supplied the fact; the inconsistency is the conjunction of the two.

Two things in this example are worth drawing out. First, the flag is available \emph{at the fourth prefix} --- the moment \emph{Zeus} is combined with \emph{The Aztec god} --- not at the end of the sentence. It is the incremental, type-driven character of CCG that makes this early flagging possible: because every prefix carries a typed, semantically interpreted category, the predication $\mathrm{aztec}(\mathrm{zeus})$ becomes available for content-checking as soon as it is composed, before the predicate and object are even read. A non-incremental parser would lift the completed sentence and query afterwards; CCG lets the query fire mid-generation, which in the synchronous regime of \cref{sec:synchronous} is the difference between detecting an error and preventing its completion. The content check requires an external source, but the \emph{timing} of the check --- its availability at the earliest structurally meaningful point --- is a contribution of the CCG layer that no mere superposition of a parser and a knowledge base would deliver.

Second, and more speculatively, the example invites a shift in how we regard the types themselves. In standard CCG a category such as $\mathit{NP}$ or $N/N$ is a combinatory instruction: it determines how an expression composes with its neighbors and what lambda term results. Here the type of a composed predication does something further. The compositional structure isolates a predication --- $\mathrm{aztec}(\mathrm{zeus})$ --- and the type of that predication determines what kind of external source could adjudicate it: a one-place property over a named entity is the kind of thing a lexical ontology can be queried about. In this sense the type functions not only as a combinatory instruction but as an interface to content-level checking: it specifies not just how the expression combines, but what external knowledge bears on its correctness and how that knowledge is to be addressed. This is the natural point of unification between the two extensions this section has developed --- the move beyond natural language and the plugging-in of external sources --- since what travels across both is the type. The reframing is not without formal kin: in the theory of higher-order contracts in programming languages \citep{findler-felleisen2002}, behavioral specifications attach to typed interfaces and \emph{blame assignment} identifies which party violated which obligation --- exactly the discipline an architecture of checks against external sources will need when a check fails and responsibility must be located. We offer this as an interpretive reframing rather than a formal result: a worked-out semantics of types-as-interfaces, and its interaction with the combinatory rules, is beyond the present paper, and we note it as a direction the framework opens rather than one it closes.
 
\subsection{The contract-and-oracle pattern}
\label{sec:foundations-oracle}
 
This substrate-versus-content distinction has a familiar realization in deployed systems, under two names that are worth bringing into contact with each other.
 
In smart-contract design, \emph{on-chain} verification --- the checking that the blockchain network itself performs as it executes a contract --- handles what is intrinsic to the contract: the type-correctness of values, the well-formedness of state transitions, the consistency of arithmetic. Anything that requires knowledge external to the chain --- the price of an asset, the outcome of an event in the physical world --- is supplied by an \emph{oracle}: a designated external source whose output the contract trusts. The contract verifies what it can verify internally; the oracle supplies what cannot be verified from on-chain state alone. The trustworthiness of any particular oracle is a meta-question that the contract architecture does not by itself answer, but the architectural separation is clear and operational.
 
In retrieval-augmented generation, the structurally identical pattern appears in natural language. A system such as Perplexity generates text using a language model and conditions the generation on documents retrieved from a trusted corpus, with citations exposed to the user. The retrieval layer plays the same role as the oracle in the smart-contract case: it supplies what the language model cannot reliably produce from its parametric memory alone, namely accurate content about the world. RAG systems contain hallucination substantially because they ground generation in retrieved context, and the practical effectiveness of this pattern is by now well established in deployed systems.
 
The framework we propose can be seen as articulating, at the architectural level, what these two patterns have in common, and as identifying what a typed compositional lifting adds to both. In the smart-contract case, on-chain verification operates at the compositional level (typing and state-transition consistency), and the oracle supplies external content; in the RAG case, the language model operates at the production level, and the retrieval layer supplies external content. What is missing in current RAG implementations is the compositional layer: the retrieved content is fed back into generation as conditioning context, where its influence is statistical rather than compositional. The framework proposes that a typed compositional lifting of generated text would produce the structured object against which retrieved content could be checked \emph{compositionally} rather than \emph{distributionally} --- the lambda term for ``The Aztec god Zeus \ldots'' makes the binding $\textit{aztec}(\textit{zeus})$ explicit and queryable, in a way that the surface text fed back to the model as conditioning does not. The architecture is the same; the lifting is what would let the check operate at the right level.
 
This framing has a final virtue worth flagging. The trustworthiness of any particular knowledge source --- a lexical ontology, a corpus of vetted Solidity patterns, a domain knowledge base --- is a meta-question that the framework recognizes but does not resolve. The framework's claim is that, given access to a trustworthy source, the typed lifting produces the substrate against which content from that source can be checked compositionally and incrementally. The selection and curation of knowledge sources is itself a substantial undertaking and is left, appropriately, to subsequent work. One recent system takes up exactly this meta-question: Eigenius \citep{will2026eigenius} builds the oracle into an auditable, typed knowledge graph whose entries carry an epistemic grade --- declared, observed, derived, or verified --- checked at commit time, so that the trustworthiness of a source becomes a property that can be inspected and re-checked rather than assumed. In the terms of this section, it is a candidate for the typed, trusted knowledge source on which the content layer depends.
 
\subsection{Incrementality and the earliest possible flagging}
\label{sec:foundations-incrementality}
 
A final foundational point concerns when the checking happens. CCG's incrementality is not just an elegant theoretical feature aligned with the prefix-driven character of autoregressive generation; it is operationally significant for the contract-and-oracle architecture we have just described. Compositional checking can flag a structural problem as soon as the structure becomes incoherent: the moment a prefix carries a type that cannot be extended to a well-formed continuation, the lifting fails. Content checking, given access to the relevant knowledge source, can flag a content problem as soon as the lifted prefix carries enough structure to be queried against the source.
 
Consider the sentence \emph{``The Aztec god Zeus \ldots''} unfolding token by token. After \emph{The}, no content check is yet evaluable. After \emph{The Aztec god Zeus}, the lifted prefix has the binding $\textit{aztec}(z) \wedge \textit{god}(z) \wedge z = \textit{zeus}$, which is queryable against a lexical ontology that records pantheon membership. At this moment --- before the sentence's predicate has even begun --- the system can flag the inconsistency, and downstream behavior (re-prompting, soft constraint, hard mask in the synchronous variant) can intervene. The same logic applies in formal languages: a Solidity function that begins an external call before its state update can be flagged at the moment the external call appears in the lifted structure, against security knowledge that records the safe ordering. An OWL class expression that begins to assert subsumption of \textit{Cow} under \textit{Carnivore} can be flagged at the moment the subsumption claim becomes queryable against a generalist ontology such as WordNet.
 
What incrementality buys, in this architecture, is the earliest possible flagging. The lifting and the knowledge-source query proceed in lockstep with generation, and intervention becomes available at the first prefix at which the check is evaluable.

The Zeus example is synthetic, constructed for minimal-pair clarity. It is worth seeing the same machinery applied to a hallucination produced in the wild by a deployed system.\footnote{We owe the episode to Mark Steedman (personal communication, June 2026), who encountered it directly.} Queried for the biblical verse in which Jacob describes his brother (``Esau my brother is a hairy man,'' Genesis 27:11), a current commercial LLM failed to supply the reference and asserted instead that the brother of Esau was \emph{Elijah} --- apparently through the lexical association that the prophet Elijah, an unrelated figure from a different period, is also described as hairy (2 Kings 1:8). The error is instructive precisely because of its etiology: it is a distributional slippage along a shared surface feature, invisible to any check based on fluency or local plausibility, and the correct answer (Jacob) is a one-step lookup in any encyclopedic knowledge base. The lifting of the offending assertion proceeds prefix by prefix, in the left-to-right style of \cref{sec:example}:

\begin{center}
\begin{tabular}{lll}
\textbf{prefix} & \textbf{category} & \textbf{partial logical form} \\[2pt]
\emph{The} & $NP/N$ & $\lambda P.\, \iota x.\, P(x)$ \\
\emph{The brother} & $NP/PP_{\textit{of}}$ & $\lambda y.\, \iota x.\, \textit{brother}(x, y)$ \\
\emph{The brother of Esau} & $NP$ & $\iota x.\, \textit{brother}(x, \textit{esau})$ \\
\emph{\ldots was} & $S/NP$ & $\lambda y.\, \iota x.\, \textit{brother}(x, \textit{esau}) = y$ \\
\emph{\ldots was Elijah} & $S$ & $\iota x.\, \textit{brother}(x, \textit{esau}) = \textit{elijah}$ \\
\end{tabular}
\end{center}

Two query moments arise, and they differ in kind from the Zeus case in a way that exercises the types-as-interfaces reading of \cref{sec:foundations-two-layers}. At the third prefix, the lifted object is a definite description over a two-place relation --- an individual-denoting term, not a one-place predication --- and its type determines the query form: not a membership check, as with $\textit{pantheon}(\textit{zeus})$, but referent resolution, $\iota x.\, \textit{brother}(x, \textit{esau}) \stackrel{?}{=} \;?$, against which an encyclopedic source returns \textit{jacob}. At the final step, the equative composes the description with the asserted identity, and the conjunction of the lifted term with the retrieved referent --- $\textit{elijah} = \iota x.\,\textit{brother}(x, \textit{esau})$ against $\textit{jacob} = \iota x.\,\textit{brother}(x, \textit{esau})$, with $\textit{jacob} \neq \textit{elijah}$ --- raises the flag at the moment the word \emph{Elijah} is emitted, the earliest structurally meaningful point. As in the Zeus case, the flag is the joint product of the two layers: the lifting supplies the queryable description and the identity structure; the external source supplies the fact.

One obligation that a real deployment of this check incurs should be stated, because the real-world example brings it into focus. The predication $\textit{brother}(x, \textit{esau})$ must trigger the query only when it occurs in a veridical position --- asserted by the producer --- and not when it is embedded under negation, a question, an attitude report (``the system claimed that the brother of Esau was Elijah''), or fictional framing, where flagging would be a false positive. This gating is not an additional burden the framework must import from elsewhere; it is information the lifting already carries. The position of a predication within the lifted term --- inside or outside the scope of a negation, a question operator, or an intensional verb --- is exactly what a compositional derivation makes explicit, and it is what context-free decomposition of text into atomic claims, as practiced in atomic-fact verification methods such as FActScore \citep{min2023}, discards. A worked treatment of veridicality projection is an implementation obligation we flag rather than discharge here, but it is an obligation the compositional route is structurally equipped to meet.

This is what makes the synchronous-coupling architecture of \cref{sec:synchronous} attractive even as a future direction: it converts post-hoc detection into prevention. The conservative post-hoc framework of this paper already exhibits the same structural advantage at a coarser granularity --- it can flag at sentence boundaries or at semantic-unit boundaries rather than at token boundaries --- but the operational character is continuous between the two, and the incrementality of CCG is what makes both possible.
 
\subsection{What this section commits us to, and what it does not}
\label{sec:foundations-summary}
 
We close this section by stating its commitments cleanly. The framework is about probabilistic prefix-driven generation by LLMs and about the architecturally appropriate checking layer for such generation. It is not about typed compositional production in general. The producer is held constant --- the LLM in its prefix-driven generative profile --- and the output language varies across natural and formal cases. The architecture has two components: production by the LLM, and knowledge of appropriate use; the latter has two layers, compositional and content. The typed lifting catches compositional failures and produces the substrate against which content checks become possible; content checking itself requires external knowledge sources, whose trustworthiness is a meta-question the framework does not resolve. The contract-and-oracle pattern from smart-contract design, and retrieval-augmented generation from natural-language deployment, are recognizable instances of this architecture in operating systems; the framework proposes that a typed compositional lifting articulates what these patterns have in common and what compositional structure adds to them. The incrementality of CCG-style derivation makes this architecture deliverable at the granularity of generation itself, supporting the earliest possible flagging of both compositional and content failures.
 
What we are not committing to: any theory of mind, any claim about LLM internals, any specific implementation of a knowledge source or its curation, any guarantee that a particular knowledge source will be trustworthy or complete, and any claim about producers whose generative profile is not prefix-driven. The commitments are architectural and historically situated. They license the cross-domain extension we develop next --- across the output languages an LLM can produce --- and they constrain the kinds of claims that extension can support.
 
\section{Beyond Natural Language: Lifting Across Typed Domains}
\label{sec:beyond}
 
The framework as developed so far treats natural language as the input distribution and CCG as the typed compositional standard against which generated text is lifted. Nothing in the operational architecture, however, depends on the input being natural language. What it depends on is something more abstract: that the language produced by the LLM admits a typed compositional standard with an incremental left-to-right interpretation, and that the production process itself proceeds left-to-right. Where both conditions hold, the lifting applies.
 
This section argues that the conditions hold for a broader class of languages than the natural ones --- and in particular for the formal languages where LLM-based generation has become consequential: programming languages, description-logic and query languages, and the broad family of expressions in formal mathematics. The argument is not that these languages are linguistically similar to English. It is that the operation of lifting incrementally-produced output into a typed compositional structure has the same formal character in all of them, by virtue of an identification whose history is older than either CCG or contemporary LLMs. Before giving that identification its formal statement, it is worth naming its programmatic ancestor. The thesis that there is no important theoretical difference between natural languages and the artificial languages of logicians, and that both can be comprehended within a single mathematically precise theory, is Montague's \citep{montague1970efl}: ``English as a formal language'' was proposed in earnest, not as a metaphor, and the extension undertaken in this section is that thesis operationalized for the setting of probabilistic prefix-driven generation.
 
\subsection{The propositions-as-types correspondence}
\label{sec:beyond-curry-howard}
 
The identification in question is the propositions-as-types correspondence, also known as the Curry--Howard isomorphism. Discovered independently by Curry in the 1930s and Howard in the 1960s, and recently surveyed at length by \citet{wadler2015}, it observes that propositions in intuitionistic logic correspond exactly to types in the simply-typed lambda calculus, and that proofs of propositions correspond exactly to programs of the corresponding type. The connective $\to$ in logic is the function-type constructor $\to$ in programming; modus ponens is function application; the construction of a proof is the construction of a program. The same correspondence has been extended to predicate logic (dependent types), second-order logic (System F and parametric polymorphism), and linear logic (session types). It is, in Wadler's phrase, ``a notion with depth, breadth, and mystery.''
 
The relevance to categorial grammar is direct, and was established early. Beginning with \citet{vanbenthem1983}, derivations in Lambek-style categorial calculi were shown to admit a Curry--Howard interpretation: a categorial proof of a sentence's grammaticality is also a lambda-calculus construction of its meaning, with the syntactic category playing the role of the type. This interpretation became foundational to the type-logical grammar tradition developed by Moortgat, Morrill, and others, where the syntax--semantics interface is explicitly framed as the Curry--Howard correspondence at work in natural language \citep{moortgat1997,morrill1994}. CCG itself has been given a logical characterization in the same family: \citet{kubota2014} shows that CCG can be understood as a substructural logic in which the directional slashes correspond to substructural implications and the combinatory rules correspond to particular combinator types.
 
What this means for the present paper is that the central operation of the framework --- assigning a typed category and a partial lambda term to a prefix, and updating both incrementally as new material arrives --- is an instance of an operation that has been studied for sixty years in mathematical logic and forty years in computer science, where it goes by various names (type-checking, type inference, proof reconstruction) but always refers to the same family of procedures. The lifting we have been calling ``CCG lifting of LLM output'' is, viewed through the Curry--Howard lens, a particular case of a more general operation: \emph{incremental type-checking of left-to-right generated content against a typed compositional standard}.
 
\subsection{Implications for cross-domain extension}
 
Once stated in this form, the framework's potential scope becomes apparent. The same operation that lifts ``A wise lord must keep faith \ldots'' into a typed lambda term lifts a program in any typed programming language into the typed term of its semantics, lifts a description-logic expression into a typed term in its ambient logic, and lifts a mathematical expression into a typed computation. In each case, the input is incrementally produced, prefix by prefix, by an LLM. In each case, a typed compositional standard exists. In each case, the lifting recovers a structured object that exposes the compositional dependencies of the generated content and makes them available for symbolic processing.
 
What changes across domains is not the operation but the type system and the lexicon. The CCG types $\catS$, $\catNP$, $\catN$, and the combinatory rules of forward and backward application, composition, and type-raising are the natural-language-specific instantiation of the more general apparatus. For a programming language, the analogous instantiation would be the language's own type system --- primitive types, function types, parameterized types, and the structural rules by which expressions in the language compose. For a description logic, it would be the types of concepts and roles and the algebra of class constructors. The framework's commitment is not to any particular type system but to the architectural pattern: incremental production from a probabilistic generator, checked against an incremental compositional analyzer that produces typed lifted representations.
 
\subsection{Programming languages}
\label{sec:beyond-pl}
 
The domain in which this cross-extension has the most immediate force is the generation of source code in typed programming languages. LLM-based code generation has become ubiquitous in software development, and the failure modes are well-documented: omitted input validation, incorrect handling of edge cases, control-flow patterns that compile but do not implement the intended semantics. Surface plausibility and structural similarity to reference implementations are routinely high, while functional correctness is substantially lower. The failure is compositional: each line of generated code is locally plausible, but the global semantic structure does not cohere with the program's intended specification.
 
A Curry--Howard reading of the lifting framework suggests where the leverage lies, and also where it does not. Generated code is already a term of a typed language, so for code the lifting has no meaning to recover, as it has for prose. What it can add is a type that the language's own type system does not compute: one that encodes the function's behavioral contract --- which preconditions it enforces, which state it changes, which events it emits. If generated code can be assigned such a type, verification reduces to type comparison: the code is a valid implementation of its intended behavior if and only if its term inhabits the expected type. This is the identification that underlies the Curry--Howard correspondence and that proof assistants such as Coq and Lean have exploited for decades. What is novel here is not the verification idea but the division of labor it implies. On the code side, where meaning is explicit, the delivered type is assigned by the language's own semantics, extended with effect-and-obligation types; as noted above, for a programming language the instantiation of the framework is the language's own type system. On the specification side, where meaning is not explicit, CCG lifts the natural-language statement of what the code should do into the expected type. And because the expected type is in hand before the code is written, the comparison can run on prefixes of the code while it is being generated, which neither testing nor compilation can do.
 
\paragraph{Solidity as a particularly clean instance.}
For readers outside the area, a brief orientation is in order. A \emph{blockchain} is a replicated, append-only ledger maintained by a decentralized network of nodes that agree, by a consensus protocol, on its contents. A \emph{smart contract} is a program stored on such a ledger and executed by the network: it holds state (for instance, account balances) and exposes functions that external parties invoke through \emph{transactions}, each of which deterministically updates the shared state. Solidity is the dominant language for writing smart contracts on Ethereum, the most widely used programming-capable blockchain, and on the many compatible platforms. Two properties make this setting consequential. Contract code, once deployed, is effectively \emph{immutable} --- it cannot be patched in place --- so a latent defect cannot simply be corrected after the fact; and contracts routinely custody economically significant assets, so a defect that is exploitable is a defect that loses money. These are the stakes against which the correctness of generated contract code must be judged.

Solidity is, for our purposes, a particularly clean case, for reasons that go beyond its relative novelty. It is less widely studied than C or Java but it embeds two features that make compositional verification both unusually tractable and unusually consequential. First, Solidity programs are explicitly \emph{transactional}: each external function call is an atomic unit of state change, either fully committed to the ledger or fully reverted, with no intermediate state observable outside the contract. This atomicity disciplines the compositional structure of the code in ways that align well with typed compositional analysis: the function boundary is also the verification boundary. Second, Solidity is \emph{event-driven} in a strong sense: contracts respond to incoming transactions by emitting \emph{events} --- typed, logged records of what occurred --- that downstream contracts and off-chain systems consume, and the type structure of those events is part of the contract's interface. The dependency between a function's input validation, its state mutations, and its emitted events is therefore explicit at the language level, in a way that makes compositional failures particularly visible to a typed lifting.
 
These features combine with high stakes to produce a domain where compositional verification of LLM-generated code is both technically tractable and economically motivated. Recent empirical work \citep{salzano2025} has shown that current LLMs --- including ChatGPT-4o, Gemini, CodeLlama, and DeepSeek-Coder --- produce Solidity smart contracts that are functionally correct only 20--26\% of the time in zero-shot generation, and at best 45\% with retrieval augmentation. The failure modes are precisely those that a typed lifting would expose: omitted \texttt{require} statements that should have validated input, missing access-control modifiers, and event emissions that do not correspond to the state changes they claim to report. Each of these is a compositional failure rather than a local one, and each is the kind of failure that surface similarity metrics will not catch.
 
We do not develop the Solidity lifting in detail here; that is a substantial undertaking, requiring an effect-and-obligation type algebra for the language, an incremental assignment of those types to contract code, and a lifting of contract specifications into the same algebra. We note only that the conceptual framework supports such a development, that the Curry--Howard correspondence provides its formal warrant, that Solidity's transactional and event-driven character makes it a particularly appropriate first target, and that the empirical case for pursuing it is strong: a method that materially improves on 20--26\% functional correctness in smart-contract generation would be a substantial contribution to a domain where deployed errors are immutable and economically consequential.

\subsubsection*{Example: A typed lifting of a Solidity function}
\label{sec:beyond-pl-example}
 
To make the type-checking-as-verification claim concrete, we apply the framework to one of the functions from the empirical study of \citet{salzano2025}. Consider \texttt{setMinDepositAmount}, a function that updates a contract's minimum-deposit parameter. The ground-truth version, drawn from a deployed contract, has three coordinated commitments: it is callable only by the owner, it validates that the new amount exceeds a minimum unit, and it emits an event recording the update. The lifted form, as a typed lambda term, captures the joint compositional structure of these commitments.
 
\begin{figure}[h]
\centering
\renewcommand{\arraystretch}{1.1}
\setlength{\tabcolsep}{1em}
\begin{tabular}{@{}p{0.46\textwidth}p{0.46\textwidth}@{}}
\toprule
\textbf{(A) Ground-truth Solidity} & \textbf{Lifted typed term and its type} \\
\midrule
\begin{minipage}[t]{\linewidth}
\vspace{0pt}
\begin{lstlisting}
function setMinDepositAmount(
  uint _amount
) external onlyOwner {
  require(_amount > unit(),
    "amount > UNIT");
  minimumDepositAmount = _amount;
  emit MinDepositUpdated(_amount);
}
\end{lstlisting}
\end{minipage}
&
\begin{minipage}[t]{\linewidth}
\vspace{0pt}\footnotesize
\(\lambda a\!:\!\textit{uint}\,\big[\textit{owner}(\textit{msg.sender})\wedge a > \textit{unit}\big].\)
\smallskip\\
\(\quad\big\langle\,\textit{set}(\textit{minDeposit},\,a),\;\textit{emit}(\textit{MinDepUpd},\,a)\,\big\rangle\)
\smallskip\\[0.6em]
\textit{of type:}
\smallskip\\
\(\quad\Pi a\!:\!\textit{uint}\,[\,\varphi_{\textit{auth}} \wedge \varphi_{\textit{val}}(a)\,].\)
\smallskip\\
\(\quad\big(\textit{State} \otimes \textit{Event}_{\textit{Upd}}\big)\)
\end{minipage}
\\
\midrule
\textbf{(B) Zero-shot LLM-generated Solidity} & \textbf{Lifted typed term and its type} \\
\midrule
\begin{minipage}[t]{\linewidth}
\vspace{0pt}
\begin{lstlisting}
function setMinDepositAmount(
  uint256 newAmount
) public onlyOwner {
  minDepositAmount = newAmount;
}
\end{lstlisting}
\end{minipage}
&
\begin{minipage}[t]{\linewidth}
\vspace{0pt}\footnotesize
\(\lambda a\!:\!\textit{uint}\,\big[\textit{owner}(\textit{msg.sender})\big].\)
\smallskip\\
\(\quad\textit{set}(\textit{minDeposit},\,a)\)
\smallskip\\[0.6em]
\textit{of type:}
\smallskip\\
\(\quad\Pi a\!:\!\textit{uint}\,[\,\varphi_{\textit{auth}}\,].\,\textit{State}\)
\end{minipage}
\\
\bottomrule
\end{tabular}
\caption{A Solidity function from \citet{salzano2025} and its hypothesized typed lifting, under both the ground-truth implementation (A) and the zero-shot LLM-generated version (B). The lifted type of (B) is a strict weakening of the lifted type of (A): the precondition omits the input-validation conjunct $\varphi_{\textit{val}}$, and the effect type collapses from a product to a single component.}
\label{fig:sol-lifting}
\end{figure}
 
In (A), the term is a dependent function over a precondition combining authorization ($\varphi_{\textit{auth}}$) and input validation ($\varphi_{\textit{val}}$), returning a product of two effects: a state mutation and an event emission. The type carries the function's behavioral contract: it states that, given an authorized caller and a validated input, the function produces both the expected state change and the expected event. In (B), the LLM-generated version has neither the validation precondition nor the event-emission effect. Its lifted type is strictly weaker than the expected contract: the precondition drops $\varphi_{\textit{val}}$, and the effect type collapses from a product to a single component.
 
The verification step is the type-comparison itself. The expected type for this function --- derived from the contract's natural-language specification or from a reference implementation --- is the one shown in (A). The LLM's output has the type shown in (B). The two are inhabitants of different types, and the latter does not inhabit the former. The compositional failure that the surface code obscures, and that similarity metrics fail to detect, is visible at the level of type structure as a strict weakening of the expected contract.
 
This is a sketch. A full development would assign such terms to generated code mechanically, rather than presenting them by hand, and would obtain type (A) from a statement of the function's intended behavior rather than from a reference implementation. It would also require a precise treatment of the refinement annotations $\varphi_{\textit{auth}}$ and $\varphi_{\textit{val}}$, which we have left informal here. What the example demonstrates is the conceptual leverage: once the lifting is in place, the failure mode documented empirically by \citet{salzano2025} --- LLM-generated contracts that compile and pass similarity checks while omitting validation and event emissions --- becomes a type-mismatch that is detectable without execution and that points directly to the omitted commitments.

The obvious objection is that type-checking is what the compiler already does. It is, but against a different type. A compiler or interpreter checks generated code against the types of its own language --- that values are \texttt{uint} or \texttt{address}, that calls match signatures --- and the defective version (B) above compiles without complaint. What no compiler can check is whether the code does what was asked, because what was asked is not in the code: it is in the request, the documentation, or the specification from which the code was generated, and Solidity's type system cannot state ``callable only by the owner'' or ``emits an update event'' as types. The lifting supplies the missing side: the expected type is obtained by lifting the specification, the delivered type by lifting the code, and the check is between the two. In the terms of \cref{sec:foundations-two-layers}, the compiler performs the compositional check; correspondence to intent is a content check that no compiler can perform from the code alone. The Navier--Stokes episode of \cref{sec:beyond-math} shows the point at its sharpest: even Lean, whose type checker is as strong as any in use, certifies that a proof compiles, not that it proves what the paper claims. In practice, the expected type is the hard part: generated code often comes with no specification beyond a brief request, and lifting such a request into a behavioral type is a harder problem than lifting the code, one we do not claim to have solved.

\paragraph{Why not testing.} A second objection is that the mismatch in \cref{fig:sol-lifting} would equally be exposed by running (B) on test cases. It would, but only on tests that exercise it: the missing validation surfaces only if some test supplies an out-of-range amount, and the missing event only if some test inspects the event log. Tests sample the input space; a type quantifies over it. Writing tests that probe the right cases, moreover, presupposes a statement of what the function ought to do, so testing does not escape the question of the expected type but inherits it. And for code that cannot be patched once deployed, a defect that testing misses is a defect that ships.

\paragraph{Where the expected type comes from.} The hard part just conceded has a partial answer in current practice. If the expected type had to be obtained from the user's informal request, the lifting would face human input, which is precisely where grammar-based parsing is weakest and where sequence-to-sequence semantic parsing --- the very mechanism that produced the code --- holds the advantage. But the request is rarely the only natural-language source: code-generating models typically state in prose what the code they are about to write will do. That statement is LLM output, the regime in which, as argued in \cref{sec:why-llm} and as the feasibility study confirms (\cref{sec:feas-coverage}), a wide-coverage grammar covers nearly every sentence. The check then needs no gold standard: does the type of the code inhabit the lifted type of the model's own description of it? The comparison is between two typed terms, not between prose and code. One could instead ask the model to state its intention directly as a typed term, but that term would be the model's own formalization, with no audit trail and exposed to the same errors as the code it is meant to check. Lifting the prose keeps the expected type tied to a statement the user can read and approve. The description need not be assumed faithful to the code: whether it is, is precisely what the check tests. What must be faithful is the description to the user's intent, and that link is checked by the user, in prose, rather than by the machine. Two limits follow. The check is only as strong as the description, since an obligation omitted from both the description and the code passes unnoticed; and the description must precede the code, since one written afterwards can simply restate what the code does. We put this forward as a direction, not a result.

\paragraph{When the check applies.} What distinguishes this arrangement from both testing and compilation, finally, is timing, and it is here that the prefix-driven character of the framework does its work. The model's description is lifted as it is produced, and on LLM prose this is not merely a theoretical property: \cref{sec:feas-availability} shows that most predications become checkable before the sentence that carries them is complete. The expected type is therefore available before the first line of code is generated. Tests need a complete, executable program, and a compiler a complete unit; against an expected type already in hand, each prefix of the code can be checked as it appears. A commitment violated within a prefix --- in the example, a state mutation reached with no validation preceding it --- can be flagged at the line where it occurs, while the function is still being generated, just as the Zeus and Esau assertions are flagged mid-sentence in \cref{sec:foundations-incrementality}.

\paragraph{Sizing the practical problem.} Since the development just called for is the natural sequel to this paper, it is worth setting out, as premises, an honest assessment of what it involves. Four observations size the undertaking.

First, the core of the undertaking is the target type algebra. Its base types are not the language's surface types --- \texttt{uint}, \texttt{address}, \texttt{mapping} --- whose correctness the compiler already polices deterministically. They are the effect-and-obligation types the example above employs: \textit{State}, \textit{Event}$_{\textit{Upd}}$, the refinement conditions $\varphi_{\textit{auth}}$ and $\varphi_{\textit{val}}$, and the product $\otimes$ on effects. On the code side, the substantive work is to assign these types incrementally, in lockstep with the traversal of the source. Whether that assignment is best given a categorial presentation, so that the code side and the specification side are terms of a single calculus, or computed by an incremental abstract interpretation of the code, is a design choice; the comparison requires only that both sides land in the same algebra.

Second, the specification side inherits both the strengths and the limits of the natural-language lifting. The statements to be lifted are LLM prose in a narrow technical register, where coverage should be high, but, as in \cref{sec:feas-coverage}, the binding constraint will be the semantic templates rather than the parser: phrases such as ``only the owner can call'', ``must exceed'', and ``emits'' must be mapped onto $\varphi_{\textit{auth}}$, $\varphi_{\textit{val}}$, and \textit{Event}. This is a small, closed domain vocabulary, and the public corpus of deployed, audited contracts supplies paired data at scale: documentation comments in natural language next to code whose effect types can be computed deterministically. The specification lifting can therefore be developed and evaluated against ground truth that natural language, where gold meanings require human annotation, has never enjoyed.

Third, the two sides must be bound to each other. The specification speaks of ``the minimum deposit''; the code names \texttt{minDepositAmount}. Generated contracts are dense with such user-defined identifiers --- function names, state variables, custom events --- that are unseen by construction. Here the program helps: the contract's own declarations type its identifiers --- the source itself states that a given name denotes an event with a \texttt{uint} parameter --- so the binding is between a typed name and a description, not between two untyped strings. It is the same vocabulary-binding problem discussed for formal mathematics in \cref{sec:beyond-math}, in a more favorable setting.

Fourth, and this is the boundary to state candidly: the properties that fall out cleanly and incrementally are the intra-function ones. Missing validation, missing event emission, the effect-type weakening of the example above --- these are visible at the function boundary, which Solidity's transactional atomicity makes coincide with the verification boundary. The properties the security community cares most about, however, with reentrancy foremost among them, are non-local: as the DAO case of \cref{sec:foundations-two-layers} illustrates, reentrancy is not visible at the external call in isolation, but only relative to whether a state update has already occurred and whether control can re-enter. The clean way to see this is that a contract is a discourse, its functions are sentences, and reentrancy is a coreference problem across sentences: the same discourse-level composition problem flagged for natural language in \cref{sec:discussion} recurs here in formal dress, and the incremental discourse machinery cited there \citep{grenander2022} is the natural candidate for lifting into this setting. Solidity is, if anything, hospitable to the move: its design carries discourse-like notions --- transactions, emitted events, inter-contract calls --- that have no counterpart in traditional languages and that map naturally onto an explicit discourse model.

These premises are set out here to size the undertaking, not to discharge it. The development itself --- the type algebra, the specification lifting, and the evaluation of the resulting checker against the empirical baseline of \citet{salzano2025} --- is the subject of work in progress.

\subsection{Description-logic and query languages}
 
The argument extends to the broad family of description-logic languages, of which two well-known instances are SQL and OWL. SQL queries generated by LLMs frequently exhibit the same failure pattern as code: schemas are misremembered, joins are constructed against absent foreign keys, aggregations are applied to incorrectly grouped relations. The relational algebra is a fully specified type system, and a categorial-style lifting could in principle catch such failures at the level of typed composition rather than at the level of execution against a database. OWL exhibits parallel failures of class subsumption, role composition, and quantifier scope: a generated class expression may be syntactically well-formed and superficially plausible while denoting a class that is unsatisfiable, equivalent to an existing class, or in an unintended subsumption relation with the surrounding ontology. The description-logic reasoning machinery already exists; what is missing is a lifting pipeline that converts LLM-generated DL expressions into the typed form that the reasoner can check.
 
Beyond SQL and OWL, the same family includes SPARQL for graph databases, the various Datalog dialects used in knowledge representation, and a growing class of domain-specific languages developed for everything from policy specification to bioinformatics. In each case, the language has an established compositional semantics, an established type-checking apparatus, and an emerging body of LLM-generated content whose compositional correctness is currently checked, when checked at all, by execution rather than by typed lifting. The framework proposed here is in principle applicable to all of these.
 
\subsection{Formal mathematics: autoformalization and the Navier--Stokes episode}
\label{sec:beyond-math}

Arithmetic we set aside in a sentence: chains of calculation whose steps do not compose into the claimed result are compositional failures a typed lifting would expose, but simpler tool-use methods (calculator APIs, program-aided generation) are usually the more practical remedy. The serious case is formal mathematics at large, and it is serious for a structural reason: here, uniquely among the domains of this section, the typed compositional standard is not something a lifting pipeline would have to construct --- it already exists in engineered form, as a proof assistant. When an LLM translates mathematical text into Lean --- \emph{autoformalization} --- it performs the mapping this paper calls lifting, from text into a typed language, but by neural translation rather than by derivation: the formal statement arrives typed, without the audit trail that a compositional lifting keeps between each part of the typed object and the stretch of text it comes from. The question the framework asks of any lifting --- what does the typed structure certify, and what does it not --- is therefore not hypothetical in this domain, and the missing audit trail is where the answer lies. It is, as of this writing, the subject of a public dispute at the highest-stakes end of mathematics, and the dispute turns on precisely the distinction this paper's \cref{sec:foundations-two-layers} draws.

In 2026, OpenAI announced a proof of finite-time blow-up of solutions to the three-dimensional Navier--Stokes equations --- a variant of a Clay Millennium Prize problem --- accompanied by a Lean~4 formalization presented as formalizing the results of the natural-language paper \citep{openai2026ns}. \citet{bastounis2026} examine whether that Lean formalization is a \emph{semantically faithful translation} of the natural-language proof, and document several respects in which it is not. Their analysis distinguishes two success criteria for autoformalization: that the produced Lean code \emph{compiles} as a proof of the formal statement, and that it \emph{faithfully translates} the natural-language argument --- its statements, and the reasoning that connects them. Their central observation is that the first criterion does not deliver the second: ``Lean acceptance (compilation) of translation does not imply semantic preservation.'' In the vocabulary of this paper, the mapping is exact. Compilation is the \emph{compositional} layer: it certifies that the formal artifact is internally coherent, that its term inhabits its stated type. Faithfulness is a \emph{content} check --- but of a distinctive kind, because the assertion being checked (``this Lean code formalizes that paper'') is located in neither artifact. It is a correspondence claim between two typed objects, and the external knowledge source that adjudicates each side is the \emph{paired artifact}: the natural-language statement is checked against the Lean declaration, and the Lean declaration against the natural-language statement. The contract-and-oracle pattern of \cref{sec:foundations-oracle} thus acquires a third instance, from mathematics itself, in which the two artifacts serve as each other's oracle.

The flagship discrepancy documented by \citet{bastounis2026} has, moreover, exactly the type-theoretic signature of our Solidity example (\cref{fig:sol-lifting}). Lemma~8.6 of the natural-language paper asserts, in its equation (8.19), an inverse estimate in which $m+4$ derivatives of the source control $m$ derivatives of the output; the Lean declaration cited for it delivers the same conclusion only under $m+5$ source derivatives. One slot differs. \Cref{fig:ns-lifting} presents the pair in the format of \cref{fig:sol-lifting}.

\begin{figure}[h]
\centering
\renewcommand{\arraystretch}{1.1}
\setlength{\tabcolsep}{1em}
\begin{tabular}{@{}p{0.46\textwidth}p{0.46\textwidth}@{}}
\toprule
\textbf{(A) NL paper, Lemma 8.6, eq.\ (8.19)} & \textbf{Lifted type (schematic)} \\
\midrule
\begin{minipage}[t]{\linewidth}
\vspace{0pt}\small
For every nonnegative integer $m$,
\[
\lVert N^{-1}F\rVert_{C_y^{m}} \;\leqslant\; C_m\,\lVert F\rVert_{C_y^{m+4}},
\]
with $N = v_t\cdot\nabla_y$ and $C_m$ independent of $F$.
\end{minipage}
&
\begin{minipage}[t]{\linewidth}
\vspace{0pt}\footnotesize
\(\Pi m\!:\!\mathbb{N}.\;\Pi F\,[\,\varphi_{\textit{src}}(F,\,m{+}4)\,].\)
\smallskip\\
\(\quad \textit{Bnd}(N^{-1}F,\,m)\)
\end{minipage}
\\
\midrule
\textbf{(B) Lean declaration cited for (8.19)} & \textbf{Lifted type (schematic)} \\
\midrule
\begin{minipage}[t]{\linewidth}
\vspace{0pt}
\begin{lstlisting}[language={},basicstyle=\ttfamily\scriptsize]
(hfirst : ... ||SmoothFourierData.xJet
   (w.length + 5) (slice f p) (x, y)||
   <= C)
\end{lstlisting}
\footnotesize (declaration \texttt{norm\_derivativeWord\_inverse\_le}; the symmetric hypothesis \texttt{hsecond} likewise requires order $m+5$)
\end{minipage}
&
\begin{minipage}[t]{\linewidth}
\vspace{0pt}\footnotesize
\(\Pi m\!:\!\mathbb{N}.\;\Pi F\,[\,\varphi_{\textit{src}}(F,\,m{+}5)\,].\)
\smallskip\\
\(\quad \textit{Bnd}(N^{-1}F,\,m)\)
\end{minipage}
\\
\bottomrule
\end{tabular}
\caption{The statement-level discrepancy in the Navier--Stokes formalization documented by \citet{bastounis2026}, presented in the format of \cref{fig:sol-lifting}. Here $\varphi_{\textit{src}}(F,k)$ abbreviates the hypothesis that the first $k$ torus derivatives of the source $F$ are uniformly bounded, and $\textit{Bnd}(G,m)$ the bounded-$C_y^m$-norm conclusion (the Lean theorem bounds each mixed derivative of order $m$ separately; as \citet{bastounis2026} note, the $C_y^m$ bound follows on taking the supremum over them). The Lean type strengthens the precondition by one derivative, so the theorem it proves is strictly weaker than the claim the paper asserts: a term of type (B) does not inhabit type (A). As in \cref{fig:sol-lifting}, the failure is invisible to compilation --- the Lean proof compiles --- and visible at the level of type comparison. The renderings are schematic and hand-constructed, as in the Solidity case; the Lean hypotheses are quoted from repository \texttt{openai/NavierStokesAndEuler} at commit \texttt{f9e8bc5}, as documented in \citet{bastounis2026}.}
\label{fig:ns-lifting}
\end{figure}

The parallel with \cref{fig:sol-lifting} is worth stating precisely, because the direction of failure differs while the signature is the same. In the Solidity case the generated artifact \emph{dropped} a precondition and an effect, delivering a weaker contract than specified; here the formal artifact \emph{strengthens} a precondition, proving a weaker theorem than asserted. In both, the delivered type and the expected type are distinct, the delivered term does not inhabit the expected type, and the discrepancy is detectable by type comparison without executing or re-deriving anything. And in both, what makes the comparison possible at all is that the two sides have been brought to typed form: the Lean side arrives typed by construction; the natural-language side must be lifted, and the lifting of a displayed norm inequality --- a $\Pi$-type over an indexed family of propositions --- is exactly the operation this paper describes, here performed by hand as in \cref{fig:sol-lifting}. Two inherited disciplines should be made explicit. First, following \citet{bastounis2026}, nothing here is a claim about the correctness of the natural-language proof itself; the claims concern the documented correspondence between the two artifacts, at the cited commit, and nothing more. Second, this case exercises the two-layer distinction and the Curry--Howard scope of the framework, not its incrementality: no one lifts a hundred-page analysis paper prefix by prefix, and the earliest-flagging argument of \cref{sec:foundations-incrementality} gets no purchase here. The unit of lifting is the statement, and the unit of checking is the statement pair.

Systems that store machine-checked proofs as evidence have begun to treat compilation as necessary but not sufficient. Eigenius \citep{will2026eigenius}, a typed knowledge-graph database with a Lean~4 checker on its commit path, grants a Lean proof \emph{verified} status only if, besides compiling, every constant in its proposition is mapped to a class of the knowledge graph. This binds the formal statement's vocabulary to the domain; it does not bind its content. A statement that requires $m+5$ derivatives where the source has $m+4$ uses exactly the same vocabulary and, on the check as described, would pass. Statement-level correspondence needs the natural-language side lifted as well, which is the step the Eigenius authors announce next: a dependently typed categorial grammar that turns the prose of a scientific argument into typed terms.

Statement-level comparison does not exhaust what can go wrong between the two artifacts, and the remaining cases show where the hard problem lives. Alongside the Navier--Stokes case, \citet{bastounis2026} give elementary examples of two further failure modes. In one, the natural-language proof is wrong, but the AI produces a different, correct Lean proof of the same statement, so compilation succeeds although the argument it was meant to verify is false. In the other, both proofs are correct but argue differently, so compilation certifies the theorem while the paper's reasoning goes unexamined. A comparison of statements cannot see either: in both, the two proofs inhabit the same theorem-type. Faithfulness of \emph{arguments} would require lifting and comparing proof structure, not statement structure --- under Curry--Howard, comparing the terms and not merely their types --- and that is a substantially harder obligation, continuous with the discourse-level composition problem flagged for natural language in \cref{sec:discussion} and for Solidity in \cref{sec:beyond-pl}.

It is at this point that a boundary result enters, and the framework should welcome rather than resist it. \citet{bastounis2026} prove that resolving the ambiguities that semantically faithful autoformalization requires is non-computable in a strong sense. Through Hilbert's tenth problem, a statement can carry an existence presupposition whose truth decides whether it can be faithfully translated at all, and resolving such ambiguities sits arbitrarily high in the Solvability Complexity Index hierarchy ($\mathrm{SCI}=\infty$), above the Halting problem ($\mathrm{SCI}=1$) --- informally, harder than any computational problem. In the vocabulary of this paper, the lifting itself remains computable --- a parser always returns its candidate readings --- but choosing among them faithfully is not: no procedure can decide, in general, under which reading, if any, a mathematical statement can be faithfully formalized. \Cref{sec:example} observed informally that genuine ambiguity leaves the lifting of a sentence a set of typed candidates rather than a single term, and obliged the architecture to state a policy; the result of \citet{bastounis2026} gives that observation theorem-grade status in the mathematical domain, where an ambiguity can encode an undecidable existence presupposition. The consequence for the framework is a precise bounding, not a refutation. What the framework supplies is a substrate on which correspondence failures, \emph{where the lifting succeeds}, become type comparisons --- as \cref{fig:ns-lifting} exhibits; what no framework can supply, by their theorem, is a procedure that always decides whether a faithful lifting exists and finds it when it does. If anything, the impossibility result strengthens the case for the two-layer discipline: since no compiler-style guarantee of faithfulness is available even in principle, explicit, auditable, type-level correspondence checks of the kind the lifting makes possible are not one option among several but the available recourse --- and their partiality is a theorem, not an engineering gap. This is the strongest form the bounded character of RQ3 takes anywhere in the paper, and we return to it in \cref{sec:discussion}.

\subsection{The unifying view}
\label{sec:beyond-unifying}

What unifies all of these domains is not their linguistic structure but their compositional structure. They are all languages in which the meaning of a complex expression is a function of the meanings of its parts and the way those parts are combined. The Curry--Howard correspondence gives this structural fact a formal expression: the meaning is the lambda term, the type is the proposition the term proves, and the lifting operation is the assignment of type and term to an incrementally produced expression. Where this combination of properties holds --- and it holds across natural language, programming languages (with Solidity as a particularly clean first target), description-logic and query languages, and formal mathematics, where the typed standard is a proof assistant and the lifting goes under the name of autoformalization --- the framework applies.

We do not claim to have demonstrated this. We have argued for the conceptual scope on the basis of the formal correspondence, indicated the empirical domain where the case for an instantiation is strongest (Solidity), and exhibited, in the Navier--Stokes episode, the two-layer distinction operating in the wild at the frontier of the mathematical literature --- together with the formal result that bounds what any such lifting can guarantee. The development of the framework in any specific formal-language domain is substantial work in its own right, and we leave it to subsequent papers, our own or others'. The point of this section is to mark the scope: the lifting framework is not specific to natural language, and the prefix-driven convergence between LLM generation and CCG-style incremental processing is the linguistic instance of a more general phenomenon whose formal articulation has been with us since Curry. Stated as a genealogy: Montague supplied the thesis --- natural language as a formal language, interpreted through a typed homomorphism; CCG supplies its incremental, surface-compositional realization; and the LLM supplies the prefix-driven producer whose output that realization can discipline.

\section{Toward Synchronous LLM--CCG Coupling}
\label{sec:synchronous}

The framework above is post hoc: it lifts text after generation. A natural and considerably more ambitious extension is to couple an incremental CCG parser to the LLM \emph{during} generation, so that at every decoding step the symbolic layer maintains a typed syntactic--semantic state aligned with the current token prefix. Let $w_{1:t}$ be the prefix at step $t$ and let the LLM produce a distribution $p(w_{t+1}\mid w_{1:t})$ over next tokens. In parallel, an incremental CCG component maintains a set of derivational states $\mathcal{S}_t$ together with partial semantic terms $\Lambda_t$. For each candidate token, the parser predicts admissible category insertions and updates these states. Information flows in both directions: the LLM proposes tokens, and the parser returns signals of admissibility, syntactic fit, and semantic consistency that may constrain or reweight generation. In a soft variant, these signals are combined with the LLM's log-probabilities; in a hard variant, inadmissible tokens are masked.

We mention this here as a future direction rather than as a fully worked-out proposal, but it is no longer purely speculative --- and for formal languages, part of the territory is already occupied. Constrained-decoding systems mask or reweight inadmissible continuations against a formal grammar during generation: PICARD does so incrementally for SQL \citep{scholak2021}, Synchromesh for several target languages via completion engines \citep{poesia2022}, and grammar-guided generation frameworks make the hard-masking regime generic and efficient \citep{willard-louf2023}. Their practical success is direct evidence for the viability of the architectural pattern. What none of them maintains, however, is a typed \emph{semantic} state --- a partial lambda term alongside the syntactic admissibility check --- and none addresses natural language, where the grammar is not given by a language definition but must be supplied by wide-coverage categorial machinery of the kind discussed above. The unoccupied territory is therefore twofold: the coupling of semantically interpreted symbolic state with generation, and its extension from formal languages to natural ones. On the CCG side, the incremental parsers of \citet{stanojevic-steedman-2019,stanojevic-steedman-2020} provide a symbolic substrate that maintains exactly the kind of typed prefix-level state $\mathcal{S}_t,\Lambda_t$ that the coupling requires, and they already use a neural supertagger as a front end --- so the architectural pattern of a neural component proposing material and a symbolic component validating and updating typed state is one for which working CCG-side machinery exists. What remains to be designed for the LLM-coupled variant is the combination rule between symbolic admissibility signals and statistical decoding scores; the operating regime in which the symbolic layer constrains rather than merely scores generation; and a treatment of pragmatic and underspecified expressions that neither tradition has fully resolved. A specific hazard of the hard variant should also be named: incremental parsers commit and reanalyze, and a mask derived from the parser's current analysis may prune continuations that a later reanalysis would have licensed --- a parser-induced garden path, in which the checking layer rather than the generator becomes the source of error. This consideration favors the soft regime, with reweighting and deferred commitment, as the default operating point for natural language, reserving hard masking for formal languages whose grammars do not give rise to reanalysis. The conservative post hoc lifting framework of the previous section remains, we believe, both more immediately defensible and more immediately useful, and we do not pretend in this draft that the synchronous version is more than a sketch.

\section{Discussion}
\label{sec:discussion}
\paragraph{The three questions, revisited.} \Cref{sec:intro} posed three questions and promised an accounting. RQ1 --- whether a typed compositional lifting of autoregressive output is well-defined --- is answered in the affirmative, and the answer is formal rather than empirical: derivational equivalence modulo composition guarantees that the lambda term does not depend on which derivation the parser follows, and the auditability of the term follows from the lexicalized character of the grammar. Nothing in this answer would change if LLMs turned out to have no internal organization resembling CCG at all; it is a fact about the grammar and the lifting, not about the model. RQ2 --- whether the lifting behaves as claimed on LLM-generated text --- receives the paper's only measured answer, and a bounded one. Derivational equivalence survives implementation: on 2,790 of 2,790 sentences where both derivations receive a well-formed logical form, a normal-form derivation and its incremental rotation are assigned the identical term (H1). Most of a sentence's checkable content is available before the sentence ends --- three quarters of composed properties, and two thirds of relations under an incremental derivation with event semantics --- and the remaining delay is principled, since a relation becomes checkable only with its last argument (H2). Coverage is limited by the templates, not the parser: with a suitable template set 97.7\% of sentences are lifted, with no difference between generating models and a modest deficit only in technical prose (H3, half confirmed). What the study establishes is that the substrate exists and is usable at the granularity the framework requires; what it does not establish is anything about the checking layers built on that substrate, which is the point at which RQ3 begins. RQ3 --- what the lifted structure adds --- we have answered at the architectural level and no further. The two-layer distinction, the reading of types as interfaces to external sources, and the earliest-flagging argument are contributions to the design of a checking architecture; they are not evidence that such an architecture, once built, outperforms retrieval-based, NLI-based, or atomic-fact methods on any benchmark, and we have been careful not to say that it does. The same holds, with greater force, for the cross-domain extension: the Curry--Howard warrant tells us the operation is the same across natural and formal languages; it tells us nothing about whether the effect-and-obligation types of a language such as Solidity can be assigned to generated code, and its specifications lifted into them, at the coverage the natural-language study achieves, and \cref{sec:beyond} sizes that undertaking rather than discharging it. The Navier--Stokes episode of \cref{sec:beyond-math} sharpens the deferral into a theorem: by the impossibility result of \citet{bastounis2026}, no procedure can decide, in general, which reading of a mathematical statement is the faithful one, so the framework's offer in that domain is bounded in principle, not merely in engineering --- type-level correspondence checks where the lifting succeeds, with the partiality of the lifting itself a formal fact rather than a gap to be closed. A reader who takes away that the paper has settled RQ1, sampled RQ2, and posed RQ3 has read it as intended.
\paragraph{What the framework claims, and what it does not.} The framework is a contribution to the conceptual architecture of neurosymbolic language processing rather than an empirical claim about the internal organization of LLMs. Its principal virtue is layered modesty: dependency grammar gives a lightweight description of observable structure, and CCG gives a reconstructive layer that supports compositional interpretation. Neither layer makes claims it cannot support. In particular, we do not claim that LLMs internally implement CCG, nor do we claim that every fluent output admits a straightforward compositional reconstruction. Ambiguity, underspecification, anaphora that cannot be resolved from local context, and pragmatic dependence all remain serious challenges, and they remain challenges for the CCG layer in the same way as for any other compositional formalizm. We would add one point of scope, developed in \cref{sec:foundations}: although the paper's entry point is the reply to a debate about natural-language grammar, the architectural claim is not confined to natural language. It concerns probabilistic prefix-driven generation as such, and applies wherever an LLM produces output in a language with a typed compositional standard --- natural or formal --- against which that output can be checked.

\paragraph{Relation to existing CCG-semantics work.} Our contribution is not the use of CCG for logical-form extraction --- that is well established in the line from Bos and Steedman through Zettlemoyer and Collins to \texttt{ccg2lambda} and its successors. Our contribution is the repositioning of this machinery in light of the convergence noted in \cref{sec:intro}: the same incremental, prefix-driven dynamic that CCG was designed to describe is the one that autoregressive language models realize statistically, which makes a CCG lifting of LLM output not merely a technical option but a structurally well-motivated one. A second point of contact is with atomic-fact verification methods such as FActScore \citep{min2023}, which decompose generated text into independent claims and check each against a source: the lifting proposed here can be read as a typed, incremental alternative to LLM-based claim decomposition --- one that preserves the entailment context of each predication (\cref{sec:foundations-incrementality}) rather than discarding it.

\paragraph{Why the convergence matters.} If a single observation deserves to be carried away from this paper, it is the one developed in \cref{sec:intro}: the prefix-driven, type-completing dynamic that autoregressive LLMs realize statistically is the dynamic that CCG was originally designed to describe. This is not, we believe, accidental. The character of left-to-right processing imposes a structure on what can be predicted and how predictions can be completed. Incrementality by itself, it should be said, is not the distinctive claim: left-corner parsers, incremental dependency parsers, and neural syntactic models all maintain prefix states of one kind or another. What CCG --- alone among major grammatical formalizms --- provides is \emph{semantically interpreted} incrementality: every prefix is a first-class typed object carrying a partial meaning, with a principled combinatory algebra governing how that object extends, rather than a parser configuration awaiting completion. It is this property, not incrementality as such, that the lifting exploits. The dependency-grammar reply to the ``grammarless'' critique is correct as far as it goes; the CCG observation extends it in a direction where what is gained is not merely a richer descriptive vocabulary but a principled route to typed compositional meaning.

\paragraph{Two layers, and a view of types.} A second strand of the paper concerns what the lifting is \emph{for}. We have argued that the typed structure it produces supports checking at two levels --- a compositional level internal to the structure, and a content level that refers the structure to an external knowledge source --- and that this division is what allows the framework to make honest contact with the problem of hallucination without overstating what compositional analysis alone can do. The same division underwrites a reading of the framework's central object that we have offered tentatively: that a type, in this setting, functions not only as a combinatory instruction but as an interface specifying what external knowledge bears on an expression's correctness and how that knowledge is to be addressed. We regard this as an interpretive reframing rather than a formal result, but it is the point at which the paper's two extensions --- the move beyond natural language and the plugging-in of external sources --- meet, since what travels across both is the type.

\paragraph{Limitations and next steps.} Four limitations of the present draft should be flagged explicitly. First, while we have provided a sentence-level example of incremental lifting, a fully worked end-to-end implementation taking a multi-paragraph passage of generated text through a CCG parser pipeline remains a natural next step. Second, the treatment of multi-sentence discourse, anaphora, and pragmatic context remains open; the lifting as described operates sentence by sentence, and the question of how to compose successive lifted representations into a coherent discourse-level object is substantial. Coreference resolution is the most acute instance of this gap; it is, however, no longer unaddressed territory. Sentence-incremental neural coreference resolution over an explicit discourse model has been developed by Grenander, Cohen, and Steedman \citep{grenander2022}, with a fuller treatment in \citet{grenander2025}, and the incrementality of that line is precisely the property a prefix-driven lifting requires of its discourse layer. Composing such a layer with the sentence-level lifting remains substantial work, but it is work for which incremental machinery now exists. Third, the robustness of available CCG parsers must be assessed accurately --- which means neither overstating nor understating it. Wide-coverage CCG parsing has historically been trained on CCGbank, the CCG conversion of the Wall Street Journal portion of the Penn Treebank, which provides roughly a million words of newswire English, while LLM-generated text ranges over registers, domains, and degrees of formality well beyond this distribution. The lexical dimension of this gap, however, is substantially mitigated by the supertagger generalization discussed in \cref{sec:bg-ccg}: modern supertaggers predict categories for unseen words from context, with the semantic category supplied by rule, and this is what allowed wide-coverage CCG parsing to be ported successfully to domains as lexically remote as biomedical text \citep{rimell-clark-2009}. The residual concern is therefore structural and register-level rather than lexical --- constructions, discourse conventions, and degrees of informality unattested in newswire --- and it is most pressing in the synchronous regime of \cref{sec:synchronous}, where parser failures would directly constrain generation rather than merely producing an unlifted sentence. Fourth, what one does with the resulting lambda terms --- whether they are used for verification, for hallucination detection, for explanation, for the extraction of decision-relevant rules, or for the construction of higher-level reasoning objects --- is held back for separate treatment. A companion paper, in preparation, details one such downstream use.

\section{Conclusion}
\label{sec:conclusion}

We have laid the groundwork for a layered perspective on grammar and meaning in large language models. Our starting point was dependency grammar, which plausibly explains important aspects of the observable grammatical organization of LLM outputs, countering the claim that such systems are entirely grammarless. We have argued that Combinatory Categorial Grammar goes further: by virtue of its prefix-driven, type-completing design, it aligns with the operational character of autoregressive generation in a way no other major grammatical formalizm does, and it supports a principled lifting of generated text into typed compositional structures. The path from dependency to compositionality is, in this sense, a path from observable fluency to recoverable meaning.

The perspective this opens brings further scenarios and challenges into view. On the one hand, the type-checking discipline that CCG naturally imposes on LLM output raises the question of how far it can actually go in catching hallucinations and the other production defects characteristic of these models. The answer we have proposed is that it can go a substantial distance, but only once the notion of ``type'' is broadened from a purely syntactic category into a channel, or interface, toward external sources of knowledge --- a move that is unsurprising and even fitting, since it mirrors the evolution of types in their native setting of programming languages, and so closes a circle with their use in the adjacent setting of natural-language analysis. On the other hand, it is precisely this move that points to the applicability of the approach across the full range of LLM output, of which natural-language text is only a part: programming code of various kinds, structured expressions in domain languages such as SQL and OWL, and the formal languages of logic and mathematics. As a consequence, the account is de-anchored from the mind-based theories of language production to which dependency grammar is inevitably tied.

Beneath this apparent proliferation of directions lies a single picture. The neural and symbolic layers of linguistic production rest on a common foundation, the one used creatively, the other normatively; and it is their pairing --- generation disciplined by compositional checking --- that the framework is meant to articulate. If the first phase of generative AI was one of rapid and largely uncontrolled production, the contribution we have sketched belongs to a second and more stable phase, in which what is generated can also be checked. This second phase, moreover, is unlikely to be navigated by any component in isolation. The threats that generative AI poses and the opportunities it opens call, respectively, for control and for enhancement, and both are most effectively pursued in a systemic setting --- through the interaction of agents, human and artificial, coordinated within shared communication spaces of the kind formalized in \citet{borghoff2025}. Within such an architecture, a CCG lifting component has a natural role to play: alongside its implications for theoretical linguistics, it contributes precisely the auditing and normative capabilities --- typed, incremental, and compositional --- that governed generation requires.

\section*{Acknowledgments}
This paper resumes a line of inquiry that the author owes to early and recent conversations with Mark Steedman, whose comments on successive drafts shaped the present version in several substantive respects: the accurate statement of supertagger generalization to unseen vocabulary, the pointer to sentence-incremental coreference resolution, and the real-world hallucination example of \cref{sec:foundations-incrementality}, which he encountered directly and generously supplied. His reading of the feasibility study, including the use of tree rotation as post-processing, and his challenge to the treatment of generated code, which led to the division of labor now adopted in \cref{sec:beyond-pl}, improved the paper further. The author is also grateful to Milo\v{s} Stanojevi\'{c} for comments on the rotation rules, including the condition on R2 and the treatment of punctuation. Errors and excesses are the author's own.

\section*{Statements and Declarations}

\paragraph{Funding.} This work was partially supported by the project Ermete (``Extended Reality for MaintenancE and Training Ecosystem''), funded by the Italian Ministry of Enterprises and Made in Italy (MIMIT), 2023, Grant number B39J25000400005.

\paragraph{Competing interests.} The author is co-founder and shareholder of BB-Smile S.r.l., a university spin-off active in blockchain and tokenization applications; the present paper does not evaluate, employ, or promote any product or service of that company. The author has no other relevant financial or non-financial interests to disclose.

\paragraph{Data availability.} Code, prompts, the generated corpus, all parses and per-sentence results, and the study protocol with its amendments are openly available at \url{https://github.com/Aribertus/ccg-lifting-feasibility}; the version described in this paper, including the full parse cache, is archived on Zenodo at \url{https://doi.org/10.5281/zenodo.24745387}.

\paragraph{Use of AI-assisted technologies.} In preparing this manuscript the author made use of Claude (Anthropic) as an assistive tool for drafting, revision, literature positioning, and LaTeX preparation, and, for the feasibility study of \cref{sec:feasibility}, for writing the study's code and running its analyses, under the author's direction and with the author's review of design, code, and results. A Claude model was also one of the four generating models whose output the study analyses; the study measures properties of the CCG instrument on that output, not the quality of any model. The author takes full responsibility for the content of the manuscript.

\bibliographystyle{plainnat}

\end{document}